\documentclass[conference]{IEEEtran}
\usepackage{times}

\usepackage[numbers]{natbib}
\usepackage{multicol}
\usepackage[bookmarks=true]{hyperref}

\let\labelindent\relax %
\usepackage[shortlabels]{enumitem}
\usepackage{wrapfig,lipsum,booktabs}
\usepackage{graphicx}
\usepackage[font=footnotesize,labelfont=bf]{caption}
\usepackage{amsmath} 
\usepackage{url}
\usepackage{seqsplit}
\usepackage{hyperref}
\hypersetup{
	colorlinks=true,
	linkcolor=orange,
	filecolor=magenta,      
	urlcolor=orange,
	citecolor=orange,
}
\usepackage{colortbl}

\usepackage{xcolor}

\begin{document}

\urlstyle{tt}  %

\title{Fine-Tuning Vision-Language-Action Models: Optimizing Speed and Success}

\newcommand{\website}{\url{https://openvla-oft.github.io}}

\author{
  \textbf{Moo Jin Kim}$^{1}$ \;\; \textbf{Chelsea Finn}$^{1}$ \;\; \textbf{Percy Liang}$^{1}$\\\\
  \large\website
}

\maketitle

\begin{abstract}
Recent vision-language-action models (VLAs) build upon pretrained vision-language models and leverage diverse robot datasets to demonstrate strong task execution, language following ability, and semantic generalization. Despite these successes, VLAs struggle with novel robot setups and require fine-tuning to achieve good performance, yet how to most effectively fine-tune them is unclear given many possible strategies. In this work, we study key VLA adaptation design choices such as different action decoding schemes, action representations, and learning objectives for fine-tuning, using OpenVLA as our representative base model. Our empirical analysis informs an Optimized Fine-Tuning (OFT) recipe that integrates parallel decoding, action chunking, a continuous action representation, and a simple L1 regression-based learning objective to altogether improve inference efficiency, policy performance, and flexibility in the model's input-output specifications. We propose OpenVLA-OFT, an instantiation of this recipe, which sets a new state of the art on the LIBERO simulation benchmark, significantly boosting OpenVLA's average success rate across four task suites from 76.5\% to 97.1\% while increasing action generation throughput by 26$\times$. In real-world evaluations, our fine-tuning recipe enables OpenVLA to successfully execute dexterous, high-frequency control tasks on a bimanual ALOHA robot and outperform other VLAs ($\pi_0$ and RDT-1B) fine-tuned using their default recipes, as well as strong imitation learning policies trained from scratch (Diffusion Policy and ACT) by up to 15\% (absolute) in average success rate.
We release code for OFT and pretrained model checkpoints at \website.
\end{abstract}

\IEEEpeerreviewmaketitle

\begin{figure*}[t]
    \centering
    \includegraphics[width=0.9\linewidth]{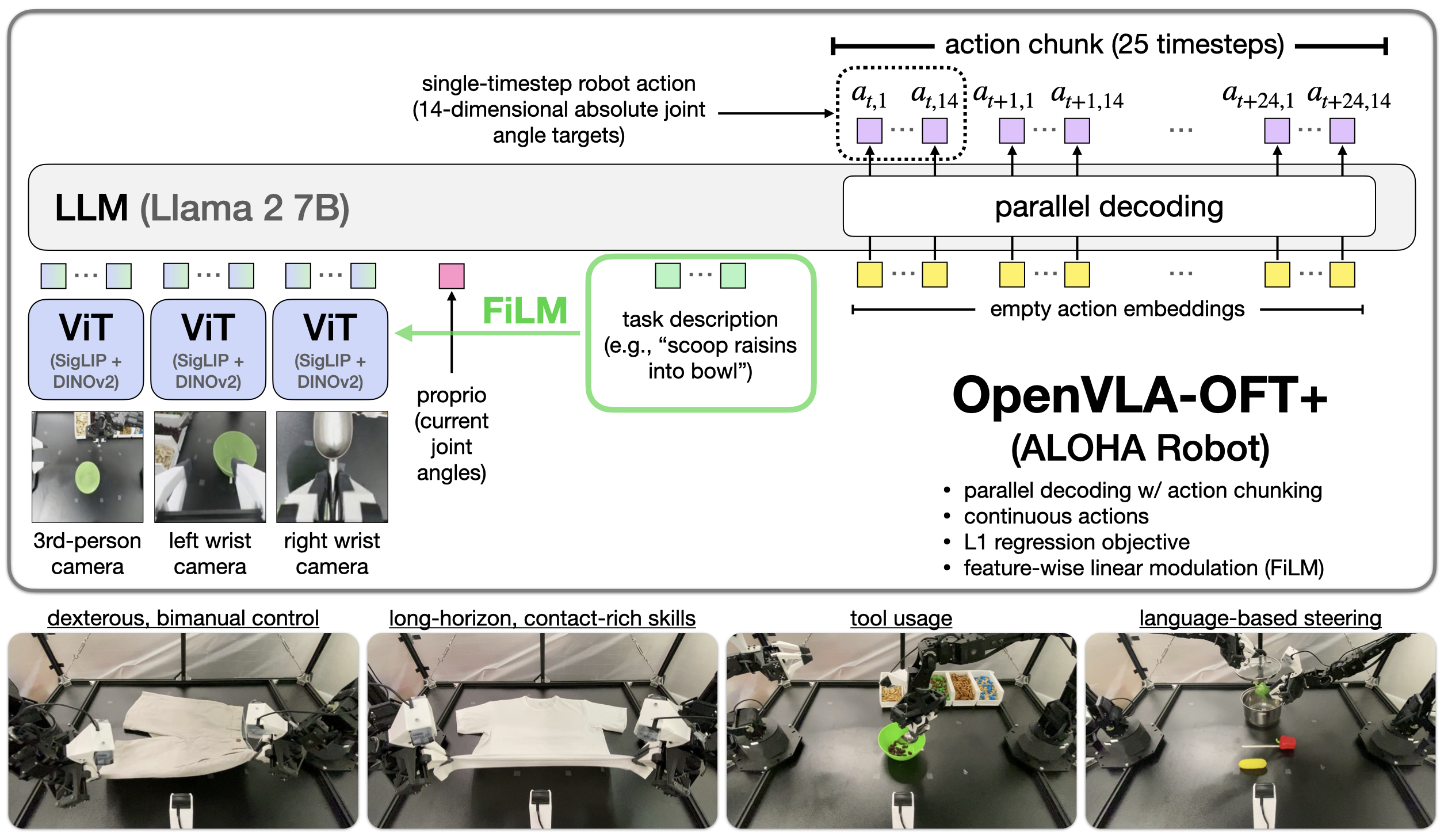}
    \caption{\textbf{OpenVLA-OFT+ on the bimanual ALOHA robot.} Our Optimized Fine-Tuning (OFT) recipe enhances fine-tuned OpenVLA policies through improved inference efficiency, model quality, and input-output flexibility. The resulting OpenVLA-OFT+ policies execute diverse dexterous manipulation tasks on a real-world bimanual robot at high control frequencies (25 Hz). The ``+'' suffix indicates the integration of feature-wise linear modulation (FiLM) \citep{perez2018film}, which strengthens language grounding in tasks where accurate language understanding is critical for success.}
    \label{fig:openvla_aloha}
\end{figure*}

\renewcommand*{\thefootnote}{\fnsymbol{footnote}}
\footnotetext{
    \raggedright
    $^1$Stanford University.
    Correspondence to:
    Moo Jin Kim \href{mailto:moojink@cs.stanford.edu}{\texttt{\seqsplit{<moojink@cs.stanford.edu>}}},
    Chelsea Finn \href{mailto:cbfinn@cs.stanford.edu}{\texttt{\seqsplit{<cbfinn@cs.stanford.edu>}}},
    Percy Liang \href{mailto:pliang@cs.stanford.edu}{\texttt{\seqsplit{<pliang@cs.stanford.edu>}}}.
}

\section{Introduction}

Recent vision-language-action models (VLAs)---robot policies built by fine-tuning pretrained vision-language models on large-scale robot datasets for low-level robotic control---have demonstrated strong task performance, semantic generalization, and language following abilities across diverse robots and tasks \citep{brohan2023rt, o2023open, li2023vision, kim2024openvla, durante2024interactive, huang2023embodied, covariant_ai_2024,wayve_ai_2024,zhen20243dvla, black2024pi_0}. Despite their strengths, fine-tuning is crucial for satisfactory deployment of VLAs on novel robots and tasks, yet it is unclear what the most effective approach for adaptation is given the large design space. A robotics practitioner who wishes to fine-tune a VLA to a new robot setup and task may default to using the same training recipe used for pretraining (or a parameter-efficient variant), but it is not obvious whether this would yield the best policy, and there is limited empirical analysis of alternative fine-tuning approaches in the literature.

Prior work has begun exploring VLA adaptation strategies, with \citet{kim2024openvla} proposing parameter-efficient fine-tuning via LoRA. However, their autoregressive action generation remains too slow (3-5 Hz) for high-frequency control (25-50+ Hz), and both LoRA and full fine-tuning of autoregressive VLAs often yield unsatisfactory performance in bimanual manipulation tasks \citep{wen2024tinyvla, liu2024rdt, black2024pi_0}. While recent approaches improve efficiency through better action tokenization schemes \citep{belkhale2024minivla, pertsch2025fastefficientactiontokenization}, achieving 2 to 13$\times$ speedups, significant latency between action chunks (e.g., 750 ms for the recent FAST approach \citep{pertsch2025fastefficientactiontokenization}) still limits real-time deployment on high-frequency bimanual robots. Exploring alternative VLA adaptation approaches that achieve both satisfactory speed and quality remains an underexplored area of research.

In this work, we study key design decisions for adapting VLAs to novel robots and tasks using OpenVLA, a representative autoregressive VLA, as our base model. We examine three key design choices: action decoding scheme (autoregressive vs. parallel generation), action representation (discrete vs. continuous), and learning objective (next-token prediction vs. L1 regression vs. diffusion). Our study reveals several key insights that build on each other: (1) parallel decoding with action chunking not only boosts inference efficiency but also improves success rates on downstream tasks while enabling greater flexibility in the model's input-output specifications; (2) continuous action representations further improve model quality compared to discrete representations; and (3) fine-tuning the VLA with an L1 regression objective yields comparable performance to diffusion-based fine-tuning while offering faster training convergence and inference speed.

Building on these insights, we introduce \textbf{OpenVLA-OFT}: an instantiation of an Optimized Fine-Tuning (OFT) recipe that integrates parallel decoding and action chunking, continuous action representations, and an L1 regression objective to enhance inference efficiency, task performance, and model input-output flexibility while maintaining algorithmic simplicity. We conduct experiments on both the standardized LIBERO simulation benchmark and dexterous tasks on a real bimanual ALOHA robot. In LIBERO, OpenVLA-OFT establishes a new state of the art by achieving 97.1\% average success rate across four task suites, outperforming both fine-tuned OpenVLA policies \citep{kim2024openvla} (76.5\%) and $\pi_0$ policies \citep{black2024pi_0} (94.2\%) while achieving a 26$\times$ speedup in action generation with 8-step action chunks. For real-world ALOHA tasks \citep{zhao2023learning}, we augment our recipe with FiLM \citep{perez2018film} for enhanced language grounding, denoting the augmented recipe as \textbf{OFT+}. OpenVLA-OFT+ successfully executes dexterous bimanual tasks like folding clothes and manipulating target food items based on the user's prompt (see Figure \ref{fig:openvla_aloha}), outperforming both fine-tuned VLAs ($\pi_0$ and RDT-1B) and prominent imitation learning policies trained from scratch (Diffusion Policy and ACT) by up to 15\% (absolute) in average success rate. With 25-timestep action chunks, OpenVLA-OFT+ achieves 43$\times$ faster throughput than base OpenVLA, demonstrating that our new fine-tuning recipe enables real-time robot control with strong task performance and language following ability.

\begin{figure*}[t]
    \centering
    \includegraphics[width=\linewidth]{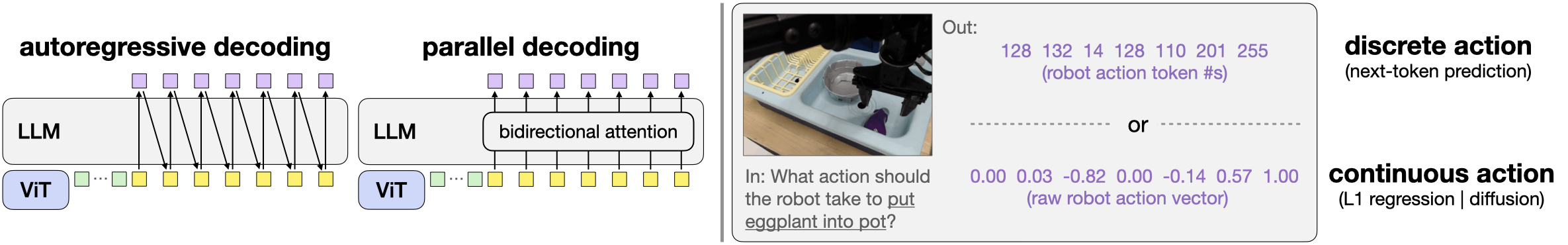}
    \caption{\textbf{Key design decisions for VLA fine-tuning.} \textbf{Left}: Comparison between autoregressive decoding, which generates actions sequentially, and parallel decoding, which leverages bidirectional attention and generates all actions in a single forward pass. \textbf{Right}: Comparison between discrete action tokens with next-token prediction and continuous action values with L1 regression or diffusion modeling objectives. The original OpenVLA training scheme includes autoregressive decoding, discrete actions, and next-token prediction.
    }
    \label{fig:ar_vs_pd_and_cont_vs_discr}
\end{figure*}

\section{Related Work}
\label{sec:related_work}

Prior works have leveraged language and vision foundation models to enhance robotic capabilities, using them as pretrained visual representations that accelerate robotic policy learning~\citep{ma2022vip, nair2022r3m, ma2023liv, Karamcheti2023LanguageDrivenRL, majumdar2023we}, for object localization in robotics tasks~\citep{gadre2023cows, stone2023open}, and for high-level planning and reasoning \citep{ahn2022icanisay, huang2022innermonologueembodiedreasoning,  singh2022progpromptgeneratingsituatedrobot, huang2022languagemodelszeroshotplanners, song2023llmplannerfewshotgroundedplanning, huang2023voxposer, duan2024manipulate}. More recently, researchers have explored fine-tuning vision-language models (VLMs) to directly predict low-level robotic control actions, producing ``vision-language-action'' models (VLAs)~\citep{brohan2023rt, o2023open, li2023vision, kim2024openvla, durante2024interactive, huang2023embodied, covariant_ai_2024,wayve_ai_2024,zhen20243dvla, wen2024tinyvla, black2024pi_0, belkhale2024minivla}, which have demonstrated effective language following \citep{zhou2023language} and generalization to out-of-distribution test conditions and unseen semantic concepts. These works focus primarily on model development, while we focus on developing a recipe for fine-tuning such models, justifying individual design decisions with insights that we gain from our empirical analysis.

Despite the importance of fine-tuning for real-world VLA deployment, empirical analysis of effective fine-tuning recipes remains limited. While \citet{kim2024openvla} study various parameter update strategies and from their findings show that LoRA fine-tuning enables effective adaptation to single-arm robots operating at low control frequencies ($<10$ Hz), their analysis does not extend to bimanual robots with high control frequencies (25-50+ Hz), a more complex control scenario. We address this gap by exploring VLA adaptation design decisions for fast inference and reliable task execution on a real-world bimanual manipulator with a 25 Hz controller.

Recent works by \citet{belkhale2024minivla} and \citet{pertsch2025fastefficientactiontokenization} improve VLA efficiency through new action tokenization schemes, using vector quantization or discrete cosine transform-based compression to represent action chunks (sequences of actions) with fewer tokens than simple per-dimension binning (as used in RT-2 \cite{brohan2023rt} and OpenVLA \citep{kim2024openvla}). While these approaches achieve 2 to 13$\times$ speedups for autoregressive VLAs, we explore design decisions beyond autoregressive modeling, which remains inherently limited by iterative generation. Our parallel decoding approach, when paired with action chunking, achieves significantly greater speedups: 26$\times$ to 43$\times$ throughput with much lower latency (0.07 ms for single-arm tasks with one input image and 0.321 ms for bimanual tasks with three input images).

Another line of research \citep{wen2024tinyvla, liu2024rdt, black2024pi_0} demonstrates effective VLA fine-tuning for high-frequency, bimanual manipulation using generative approaches like diffusion or flow matching. While these diffusion-based VLAs achieve higher action throughput than autoregressive VLAs by generating multi-timestep action chunks simultaneously, they introduce computational trade-offs through slower training and multiple denoising or integration steps at inference time. Furthermore, these diffusion VLAs vary considerably in architecture, learning algorithm, vision-language fusion approach, and input-output specifications---and which design elements most significantly impact performance remains unclear. Through controlled experiments, we show that policies fine-tuned with a simpler L1 regression objective can match more complex approaches in task performance while achieving significantly greater inference efficiency.

Lastly, unlike prior work that relies on a separate faster low-level control policy in addition to a slower VLA \citep{pan2024vision}, in this work we fine-tune the base VLA end-to-end with techniques that enable high efficiency without needing a separate controller. Further, unlike other works that collect online interaction data to continually update specialist policies via reinforcement learning \citep{julg2025refined}, here we focus on the simpler imitation learning paradigm in which we develop high-performing policies just by training once offline on a fixed dataset of expert task demonstrations.

\section{Preliminaries}
\label{sec:preliminaries}

\textbf{Original OpenVLA formulation.} We use OpenVLA \citep{kim2024openvla} as our representative base VLA, a 7B-parameter manipulation policy created by fine-tuning the Prismatic VLM \citep{karamcheti2024prismatic} on 1M episodes from the Open X-Embodiment dataset \citep{o2023open}. See Appendix \ref{app:model_architecture_details} for architecture details. OpenVLA's original training formulation uses \textbf{autoregressive} prediction of 7 \textbf{discrete} robot action tokens per timestep: 3 for position control, 3 for orientation control, and 1 for gripper control. It employs \textbf{next-token prediction} with cross-entropy loss as its learning objective, similar to language models. We explore alternative formulations including parallel decoding, continuous action representations, and learning objectives like L1 regression and diffusion modeling in the next few sections.

\textbf{Action chunking.} Prior works have shown that action chunking---i.e., predicting and executing a sequence of future actions without intermediate replanning---improves policy success rates across many manipulation tasks \citep{zhao2023learning, chi2023diffusion, liu2024bidirectional}. However, OpenVLA's autoregressive generation scheme makes action chunking impractical, as generating even a single-timestep action takes 0.33 seconds on an NVIDIA A100 GPU. For a chunk size of $K$ timesteps and action dimensionality $D$, OpenVLA requires $KD$ sequential decoder forward passes versus just $D$ passes without chunking. This $K$-fold increase in latency makes action chunking impractical for high-frequency robots under the original formulation. In the next section, we present a parallel generation scheme that enables efficient action chunking.

\section{Studying Key VLA Fine-Tuning Design Decisions}
\label{sec:design_decisions}

In this section, we first outline key design decisions for adapting VLAs to novel robot setups and tasks and provide details on their implementation.

\subsection{VLA Fine-Tuning Design Decisions}
\label{sec:finetuning_design_decisions}

Existing approaches that fine-tune VLAs using the base model's autoregressive training recipe face two key limitations: slow inference speed (3-5 Hz) unsuitable for high-frequency control, and unreliable task execution on bimanual manipulators \citep{wen2024tinyvla, liu2024rdt, black2024pi_0}.

To address these challenges, we investigate three key design components for VLA fine-tuning:
\begin{enumerate}[(a)]
    \item \textbf{Action generation strategy (Figure \ref{fig:ar_vs_pd_and_cont_vs_discr}, left):} We compare autoregressive generation, which requires sequential token-by-token processing, with parallel decoding, which generates all actions simultaneously and enables efficient action chunking.

    \item \textbf{Action representation (Figure \ref{fig:ar_vs_pd_and_cont_vs_discr}, right):} We examine discrete actions (256-bin discretization of normalized actions) processed through softmax-based token prediction, versus continuous actions directly generated by an MLP action head. For discrete actions, the final hidden states of the language model decoder are linearly projected into logits, which are processed by a softmax operation to form the probability distribution over action tokens. For continuous actions, the final hidden states are instead mapped directly to normalized continuous actions by a separate action head MLP.
    
    \item \textbf{Learning objective (Figure \ref{fig:ar_vs_pd_and_cont_vs_discr}, right):} We compare policies fine-tuned with next-token prediction for discrete actions, L1 regression for continuous actions \citep{zhao2023learning}, and conditional denoising diffusion \citep{chi2023diffusion} for continous actions  (similar to \citet{chi2023diffusion}).
\end{enumerate}

We conduct our study using OpenVLA \citep{kim2024openvla} as the base model, adapting it via LoRA fine-tuning \citep{hu2021lora} due to our relatively small training datasets (500 demonstrations versus 1M demonstrations for pretraining).

\subsection{Implementing Alternative Design Components}
\label{sec:implementing_designs}

The base OpenVLA model originally employs autoregressive generation of discrete action tokens optimized via next-token prediction. We implement alternative design decisions for fine-tuning while keeping the original pretraining unchanged. We describe the key implementation aspects below, with further details explained in Appendix \ref{app:implementation_details}.

\textbf{Parallel decoding and action chunking.} Unlike autoregressive generation which requires sequential token prediction, parallel decoding enables the model to map input embeddings to the predicted output sequence in a single forward pass. We modify the model to receive empty action embeddings as input and replace the causal attention mask with bidirectional attention, allowing the decoder to predict all actions simultaneously. This reduces action generation from $D$ sequential passes to a single pass, where $D$ is the action dimensionality.

Parallel decoding naturally extends to action chunking: to predict actions for multiple future timesteps, we simply insert additional empty action embeddings in the decoder's inputs, which are then mapped to a chunk of future actions. For chunk size $K$, the model predicts $KD$ actions in one forward pass, increasing throughput $K$-fold with minimal latency impact. While parallel decoding may theoretically be less expressive than autoregressive approaches, our experiments show no performance degradation across diverse tasks.

\textbf{Continuous action representations.} OpenVLA originally uses discrete action tokens where each action dimension is normalized to $[-1,+1]$ and uniformly discretized into 256 bins. While this approach is convenient since it requires no architectural modifications to the underlying VLM, the discretization process can sacrifice fine-grained action details. We study continuous action representations with two learning objectives drawn from prominent imitation learning approaches:

First, similar to \citet{zhao2023learning}, we implement L1 regression by replacing the decoder's output embedding layer with an MLP action head that directly maps final decoder layer hidden states to continuous action values. The model is trained to minimize the mean L1 difference between predicted and ground-truth actions, maintaining the efficiency benefits of parallel decoding while potentially improving action precision.

Second, inspired by \citet{chi2023diffusion}, we implement conditional denoising diffusion modeling where the model learns to predict noise added to action samples during forward diffusion. During inference, the policy gradually denoises noisy action samples via reverse diffusion to produce real actions. While this approach offers potentially more expressive action modeling, it requires multiple forward passes during inference (50 diffusion steps in our implementation), impacting deployment latency even with parallel decoding.

\textbf{Additional model inputs and outputs.} While the original OpenVLA processes a single camera view, some robot setups include multiple viewpoints and additional robot state information. We implement a flexible input processing pipeline: For camera images, we use OpenVLA's dual vision encoder to extract 256 patch embeddings per view, which are projected into the language embedding space with a shared projector network. For low-dimensional robot state inputs (e.g., joint angles and gripper state), we employ a separate projection network to map these into the same embedding space as one additional input embedding.

All input embeddings---visual features, robot state, and language tokens---are concatenated along the sequence dimension before being passed to the decoder. This unified latent representation enables the model to attend to all available information when generating actions. Combined with parallel decoding and action chunking, this architecture can efficiently process rich multimodal inputs while generating multiple timesteps of actions, as illustrated in Figure \ref{fig:openvla_aloha}.

\subsection{Augmenting OpenVLA-OFT with FiLM for Enhanced Language Grounding.}
\label{sec:implementing_film}

\textbf{Challenges with language following.} When deploying on the ALOHA robot setup with multiple viewpoints including from wrist-mounted cameras, we observe that policies can struggle with language following due to spurious correlations in visual inputs. During training, policies may learn to latch onto such spurious correlations when predicting actions, rather than properly attending to the language instructions, resulting in poor adherence to the user's commands at test time. Furthermore, language inputs may only be critical at specific moments in a task---for example, after grasping the spoon and deciding which ingredient to scoop in the ``scoop X into bowl'' task discussed in Section \ref{sec:expts_adapting_to_aloha}. Therefore, without special techniques, training the model to appropriately focus on language inputs can be particularly challenging.

\textbf{FiLM.} To enhance language following, we employ feature-wise linear modulation (FiLM) \citep{perez2018film}, which infuses language embeddings into the visual representations so that the model pays more attention to the language inputs. We compute the average of the language embeddings $\mathbf{x}$ from the task description and project it to obtain scaling and shifting vectors $\mathbf{\gamma}$ and $\mathbf{\beta}$. These vectors modulate the visual features $\mathbf{F}$ through an affine transformation:

\[
\text{FiLM}(\mathbf{F}|\mathbf{\gamma},\mathbf{\beta}) = \mathbf{\hat{F}} = (1 + \mathbf{\gamma}) \odot \mathbf{F} + \mathbf{\beta}
\]

A crucial implementation detail is the choice of what represents a ``feature'' for modulation in vision transformers. While one might naturally consider treating individual patch embeddings as features to be modulated, we find that this approach results in poor language following. Instead, drawing from how FiLM operates in convolutional networks, where modulation applies spatially-agnostically by scaling and shifting entire feature maps, we apply each element of $\mathbf{\gamma}$ and $\mathbf{\beta}$ to the corresponding hidden unit across all visual patch embeddings so that $\mathbf{\gamma}$ and $\mathbf{\beta}$ influence all patch embeddings. Concretely, this makes $\mathbf{\gamma}$ and $\mathbf{\beta}$ $D_{ViT}$-dimensional vectors, where $D_{ViT}$ is the number of hidden dimensions (i.e., the number of elements in each of the patch embeddings in the vision transformer's latent representations).

We apply FiLM after the self-attention layer and before the feedforward layer in each vision transformer block, with separate projectors for each block (see Figure \ref{fig:film}). Additional implementation details are provided in Appendix \ref{app:film_details}. We only use FiLM for the ALOHA experiments discussed in Section \ref{sec:expts_adapting_to_aloha}, where multiple camera viewpoints lead to a larger presence of spurious correlations in visual inputs.

\section{Experiments: Evaluating VLA Fine-Tuning Design Decisions}
\label{sec:expts_evaluating_design_decisions}

\begin{figure}[t]
    \centering
    \includegraphics[width=\linewidth]{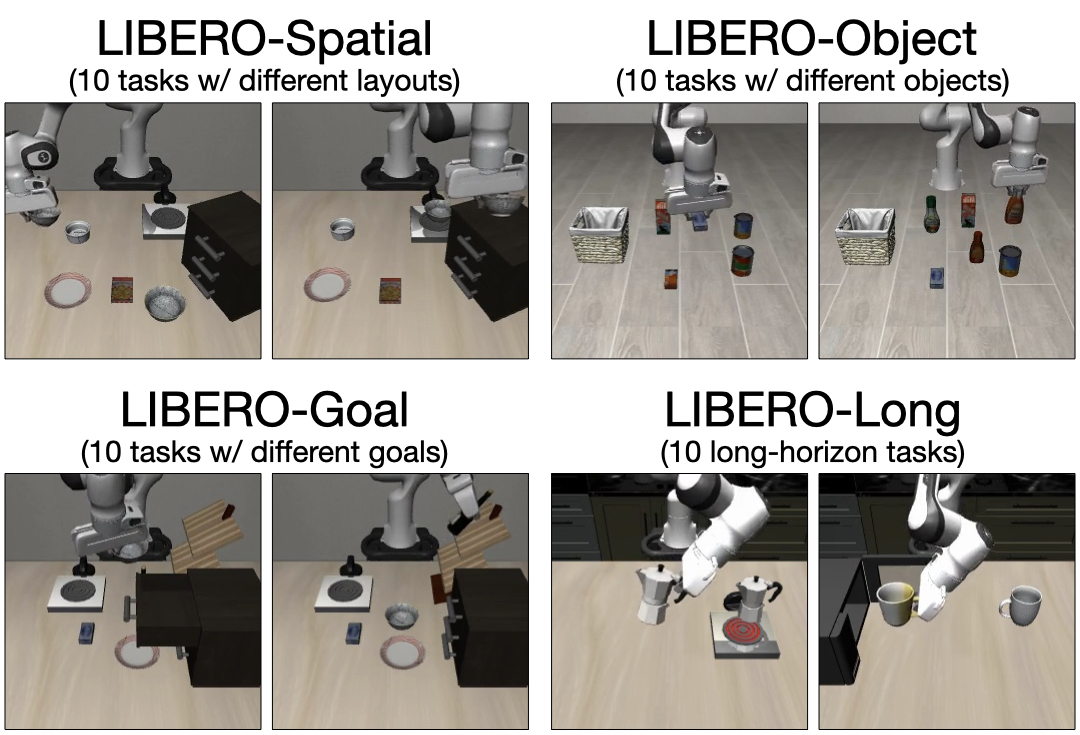}
    \caption{\textbf{LIBERO simulation benchmark \citep{liu2024libero} task suites.} We study VLA fine-tuning design decisions using four representative task suites. Here we depict two of ten tasks per suite.}
    \label{fig:libero_suites}
\end{figure}

In this section, we evaluate the effects of our proposed VLA adaptation design decisions through controlled experiments aimed at answering three key questions:

\begin{enumerate}
    \item How does each design decision affect the fine-tuned policy's success rate on downstream tasks?
    \item How does each design decision affect model inference efficiency (action generation throughput and latency)?
    \item How do the alternative fine-tuning formulations affect flexibility in model input-output specifications?
\end{enumerate}

\subsection{LIBERO Experimental Setup}

We evaluate on the LIBERO simulation benchmark \citep{liu2024libero}, which features a Franka Emika Panda arm in simulation with demonstrations containing camera images, robot state, task annotations, and delta end-effector pose actions. We use four task suites---LIBERO-Spatial, LIBERO-Object, LIBERO-Goal, and LIBERO-Long---each providing 500 expert demonstrations across 10 tasks to assess policy generalization to different spatial layouts, objects, goals, and long-horizon tasks.

Following \citet{kim2024openvla}, we filter unsuccessful demonstrations and fine-tune OpenVLA via LoRA \citep{hu2021lora} on each task suite independently. We train for 50-150K gradient steps for non-diffusion methods and 100-250K steps for diffusion methods (which converge slower), using a batch size of 64-128 across 8 A100/H100 GPUs. We test checkpoints every 50K steps and report the best performance for each run. Unless specified otherwise, policies receive one third-person image and language instruction as input. For methods using action chunking, we set chunk size to $K=8$ to match the Diffusion Policy baseline \citep{chi2023diffusion}, and execute full chunks before replanning, which we find improves both speed and performance. See Appendix \ref{app:openvla_oft_training_details} for hyperparameter details.

Our primary baseline in this study is the base OpenVLA model fine-tuned using the original fine-tuning recipe. However, for broader comparison, we also include LIBERO results from prior state-of-the-art imitation learning methods, such as Diffusion Policy \citep{chi2023diffusion}, Octo \citep{team2024octo}, DiT Policy \citep{hou2025diffusiontransformerpolicyscaling}, Seer \citep{tian2024predictive}, MDT \citep{reuss2024multimodaldiffusiontransformerlearning}, and $\pi_0$ \citep{black2024pi_0}. Note that Seer uses additional LIBERO-90 pretraining data.

\begin{table*}[t]
\centering
\caption{\textbf{LIBERO task performance results.} Success rates (SR) across LIBERO benchmark task suites \citep{liu2024libero}. Results include policies fine-tuned from pretrained base models (Octo, DiT Policy, Seer, $\pi_0$), models trained from scratch (Diffusion Policy, Seer (scratch), MDT), and OpenVLA variants using different fine-tuning design decisions: parallel decoding (PD), action chunking (AC, chunk size $K=8$ timesteps), and continuous actions with L1 regression (Cont-L1) or with diffusion (Cont-Diffusion). Unless otherwise specified, OpenVLA diffusion policies use 50 diffusion steps at both train and test time. OpenVLA results from this work are averaged over 500 trials for each task suite (10 tasks × 50 episodes). Our full OpenVLA-OFT method with additional inputs achieves state-of-the-art 97.1\% average success rate. Baseline results are from the original papers, except for Diffusion Policy, Octo, and the original OpenVLA policies, whose results are reported by \citet{kim2024openvla}. Bold and underlined values indicate best and second-best performance. We separate the comparisons based on which inputs the policies receive as well as whether or not they were trained using modified datasets following \citet{kim2024openvla} (in the modified versions, actions with near-zero magnitude and unsuccessful demonstrations are filtered out). \, *: MDT results with 100\% language annotations were obtained through direct correspondence with the authors.}
\label{tab:policy_performance_results}
\resizebox{0.8\textwidth}{!}{\begin{tabular}{l|c|c|c|c|c}
\toprule
\multicolumn{6}{c}{Policy inputs: third-person image, language instruction} \\
\multicolumn{6}{c}{(Modified training dataset; unsuccessful demonstrations filtered out)} \\
\midrule
& Spatial & Object & Goal & Long & Average \\
& SR (\%) & SR (\%) & SR (\%) & SR (\%) & SR (\%) \\
\midrule
Diffusion Policy (scratch) \citep{chi2023diffusion} & 78.3 & 92.5  & 68.3  & 50.5   & 72.4 \\
Octo (fine-tuned) \citep{team2024octo} & 78.9 & 85.7 & 84.6 & 51.1 & 75.1 \\
DiT Policy (fine-tuned) \citep{hou2025diffusiontransformerpolicyscaling} & 84.2 & 96.3 & 85.4 & 63.8 & 82.4 \\
OpenVLA (fine-tuned) \citep{kim2024openvla} & 84.7 & 88.4 & 79.2 & 53.7 & 76.5  \\
OpenVLA (fine-tuned) + PD\&AC & 91.3 & 92.7 & 90.5 & 86.5 & 90.2 \\
OpenVLA (fine-tuned) + PD\&AC, Cont-Diffusion & \textbf{96.9} & \underline{98.1} & \underline{95.5} & \textbf{91.1} & \textbf{95.4} \\
\textbf{OpenVLA-OFT} (OpenVLA (fine-tuned) + PD\&AC, Cont-L1) (ours) & \underline{96.2} & \textbf{98.3} & \textbf{96.2} & \underline{90.7} & \underline{95.3}  \\
\midrule
\multicolumn{6}{c}{Policy inputs: third-person image, wrist camera image, robot proprioceptive state (optional), language instruction} \\
\multicolumn{6}{c}{(Original unfiltered training dataset)} \\
\midrule
& Spatial & Object & Goal & Long & Average \\
& SR (\%) & SR (\%) & SR (\%) & SR (\%) & SR (\%) \\
\midrule
MDT (scratch; with 2\% language annotations) \citep{reuss2024multimodaldiffusiontransformerlearning} & \underline{78.5} & 87.5 & \underline{73.5} & 64.8 & \underline{76.1} \\
MDT (scratch; with 100\% language annotations)* & \textbf{95.2} & \textbf{97.8} & -- & 83.0 & -- \\
Seer (scratch) \citep{tian2024predictive} & -- & -- & -- & 78.7 & -- \\
Seer (pretrained on LIBERO-90, then fine-tuned) \citep{tian2024predictive} & -- & -- & -- & \underline{87.7} & -- \\
\textbf{OpenVLA-OFT} (OpenVLA (fine-tuned) + PD\&AC, Cont-L1) (ours) & \textbf{95.2} & \underline{94.2} & \textbf{95.2} & \textbf{93.2} & \textbf{94.5}  \\
\midrule
\multicolumn{6}{c}{Policy inputs: third-person image, wrist camera image, robot proprioceptive state (optional), language instruction} \\
\multicolumn{6}{c}{(Modified training dataset; unsuccessful demonstrations filtered out)} \\
\midrule
& Spatial & Object & Goal & Long & Average \\
& SR (\%) & SR (\%) & SR (\%) & SR (\%) & SR (\%) \\
\midrule
$\pi_0$ + FAST (fine-tuned) \citep{pertsch2025fastefficientactiontokenization} & 96.4 & 96.8 & 88.6 & 60.2 & 85.5 \\
$\pi_0$ (fine-tuned) \citep{black2024pi_0} & \underline{96.8} & \textbf{98.8} & \underline{95.8} & \underline{85.2} & \underline{94.2} \\
\textbf{OpenVLA-OFT} (OpenVLA (fine-tuned) + PD\&AC, Cont-L1) (ours) & \textbf{97.6} & \underline{98.4} & \textbf{97.9} & \textbf{94.5} & \textbf{97.1}  \\
\bottomrule
\end{tabular}}
\vspace{-0.2cm}
\end{table*}

\subsection{LIBERO Task Performance Comparisons}

For satisfactory deployment, robot policies must demonstrate reliable task execution. We first assess how different VLA fine-tuning design decisions affect success rates on the LIBERO benchmark.

Our efficiency analysis (which we discuss later) reveals that parallel decoding (PD) and action chunking (AC) together are necessary for high-frequency control (25-50+ Hz), especially for bimanual robots with double the amount of action dimensions. We therefore evaluate OpenVLA policies with both techniques used jointly, comparing variants using discrete actions, continuous actions with L1 regression, and continuous actions with diffusion.

Results in Table \ref{tab:policy_performance_results} show that parallel decoding and action chunking not only increase throughput but also improve performance significantly, raising average success rates by 14\% (absolute) over autoregressive OpenVLA policies. This improvement is particularly pronounced in LIBERO-Long, suggesting that action chunking helps capture temporal dependencies \citep{liu2024bidirectional} and reduce compounding errors \citep{ross2011reduction}, which ultimately leads to smoother and more reliable task execution. In addition, we find that using continuous action variants further improves success rates by 5\% (absolute) over the discrete action variant, likely due to higher precision in the action predictions. L1 regression and diffusion variants achieve comparable performance, indicating that the high-capacity OpenVLA model can effectively model the multi-task action distribution even with simple L1 regression.

\begin{table}
\caption{\textbf{LIBERO inference efficiency results.} Action generation throughput and latency for 7-dimensional actions, averaged over 100 queries on an NVIDIA A100 GPU. Each query processes a 224 × 224 px image and a LIBERO task command (``pick up the alphabet soup and place it in the basket''). We compare OpenVLA variants with parallel decoding (PD), action chunking ($K=8$ timesteps), and continuous actions using L1 regression or diffusion objectives. For the diffusion variants, we report inference efficiency measurements with varying numbers of denoising steps at test time using the DDIM sampler \citep{song2020denoising}, along with the corresponding success rates in the LIBERO-Long task suite. In the final row, we report the inference efficiency of our proposed OpenVLA-OFT formulation with additional inputs, such as wrist camera image and robot state.}
\label{tab:efficiency_results}
\resizebox{0.5\textwidth}{!}{
\begin{tabular}{lcc|c}
\toprule
& Throughput (Hz) $\uparrow$ & Latency (Sec) $\downarrow$ & LIBERO-Long SR (\%) \\
\midrule
OpenVLA & 4.2 & 0.2396 & 53.7 \\
\;\; + PD & 15.9 & \textbf{0.0629} & -- \\
\;\; + PD\&AC & 108.8 & 0.0735 & 86.5 \\
\;\; + PD\&AC, Cont-L1 & \textbf{109.7} & \underline{0.0729} & 90.7 \\
\;\; + PD\&AC, Cont-Diffusion & & \\
\;\;\;\;\;\;\;\; $T_{train}=50$, $T_{test}=50$ & 4.2 & 1.9070 & \textbf{91.1} \\
\;\;\;\;\;\;\;\; $T_{train}=50$, $T_{test}=10$ & 19.3 & 0.4145 & \underline{91.0} \\
\;\;\;\;\;\;\;\; $T_{train}=50$, $T_{test}=5$ & 35.1 & 0.2279 & 90.0 \\
\;\;\;\;\;\;\;\; $T_{train}=50$, $T_{test}=2$ & 80.3 & 0.0996 & 85.7 \\
\;\;\;\;\;\;\;\; $T_{train}=50$, $T_{test}=1$ & \underline{109.4} & 0.0731 & 0.0 \\
\midrule
\;\; + PD\&AC, Cont-L1 & & & \\
\;\;\;\; + Additional Inputs (wrist img, proprio) & 71.4 & 0.1120 & 94.5 \\
\bottomrule
\end{tabular}}
\vspace{-0.4cm}
\end{table}

\subsection{LIBERO Inference Efficiency Comparisons}

Efficient inference is crucial for deploying VLAs on high-frequency control robots. We now evaluate how parallel decoding (PD), action chunking (AC), and continuous action representations affect model inference speed. We measure average latency (time to generate one robot action or action chunk) and throughput (total actions generated per second) by querying each model variant 100 times on an NVIDIA A100 GPU. Each query processes a 224 x 224 px image and a sample LIBERO language instruction (``pick up the alphabet soup and place it in the basket'').

Results in Table \ref{tab:efficiency_results} show that parallel decoding reduces latency and increases throughput by 4$\times$ by replacing 7 sequential forward passes through the decoder portion of the policy with a single pass. Adding action chunking ($K=8$) increases latency by 17\% due to longer attention sequences in the decoder, but when combined with parallel decoding, it dramatically improves throughput, achieving a 26$\times$ speedup over baseline OpenVLA. The continuous actions variant with L1 regression shows negligible difference in efficiency since the additional MLP action head adds minimal computational cost compared to the base model. The primary diffusion variant requires 50 denoising steps (since this is the number of steps specified at train time) and thus suffers from high latency. However, it still achieves the same effective throughput as baseline OpenVLA due to parallel decoding and action chunking. This means that despite longer pauses between action chunks, the 50-step diffusion variant still completes robot episodes at the same speed as the original autoregressive variant. We also test the OpenVLA diffusion variant using less than 50 denoising steps at inference time (enabled by the DDIM sampler \citep{song2020denoising}), matching state-of-the-art diffusion and flow matching methods that use 5-10 steps at inference time \citep{reuss2024multimodaldiffusiontransformerlearning, liu2024rdt, black2024pi_0}. These configurations result in much greater inference efficiency, as shown in Table \ref{tab:efficiency_results}. However, the success rates decrease with fewer denoising steps due to reduced model quality.

\subsection{Model Input-Output Flexibility}

As explained in Section \ref{sec:implementing_designs} and validated by our efficiency evaluations in the prior section, parallel decoding enables OpenVLA to generate action chunks with minimal latency increase, thereby enhancing flexibility in model \emph{outputs}. The significant speedup realized by parallel decoding and action chunking creates headroom for processing additional model \emph{inputs} as well. We demonstrate this by fine-tuning OpenVLA with additional inputs such as robot proprioceptive state and a robot wrist-mounted camera image, which doubles the number of visual patch embeddings being passed into the language model decoder, from 256 to 512. Despite this substantial increase in input sequence length, the fine-tuned OpenVLA policy maintains high throughput (71.4 Hz) and low latency (0.112 sec), as shown in Table \ref{tab:efficiency_results}.

Evaluating these policies with additional inputs on the LIBERO benchmark reveals further improvements in average success rate across all task suites (Table \ref{tab:policy_performance_results}). Notably, our enhanced fine-tuned OpenVLA policies outperform even the best fine-tuned $\pi_0$ policies \citep{black2024pi_0, pertsch2025fastefficientactiontokenization}---which benefit from a base model with larger-scale pretraining and a more sophisticated learning objective (flow matching \citep{lipman2022flow})---as well as Multimodal Diffusion Transformer (MDT) \citep{reuss2024multimodaldiffusiontransformerlearning} and Seer \citep{tian2024predictive} policies. Even with a simpler base model pretrained on less data than more recent VLAs, we find that our alternative VLA adaptation design decisions empower fine-tuned OpenVLA policies to establish a new state of the art on the LIBERO benchmark.

\subsection{Optimized Fine-Tuning Recipe}

Based on the demonstrated improvements in task performance, inference efficiency, and model input-output flexibility, we propose an \textbf{Optimized Fine-Tuning (OFT)} recipe for VLA adaptation that combines three key components:
\begin{enumerate}
    \item parallel decoding with action chunking
    \item continuous action representation
    \item L1 regression objective
\end{enumerate}

These design choices work together to produce strong policies that can be deployed at high frequencies while maintaining algorithmic simplicity. We denote policies fine-tuned from the OpenVLA base model using our OFT recipe as \textbf{OpenVLA-OFT}. In Section \ref{sec:expts_adapting_to_aloha}, we evaluate OpenVLA-OFT's capabilities on dexterous, bimanual manipulation tasks in the real world using a high-frequency control robot.

\subsection{Additional Experiments}

Given that the alternative fine-tuning formulation, along with additional model inputs and outputs, induces a distribution shift between the base VLA's pretraining and fine-tuning, one reasonable question is whether the base VLA's pretrained representations are helpful and have any influence on the results we have reported. We conduct an ablation study in Appendix \ref{app:ablating_film_libero} to address this question, ablating the OpenVLA pretraining phase and directly fine-tuning the underlying pretrained VLM with the OFT recipe. As shown in Table \ref{tab:app:openvla_scratch_ablation_libero}, the base OpenVLA pretrained representations are indeed still beneficial for robotic policy learning, as removing them leads to a 5.2\% drop in average success rate (absolute) in our LIBERO evaluation suite.

(See Appendix \ref{app:additional_experiments} for other additional experiments, including scaling OpenVLA-OFT up to larger datasets---in both the LIBERO simulation setting as well as a real-world single-arm robot manipulation setting with over 50K demonstrations from the BridgeData V2 dataset \citep{walke2023bridgedata}.)

\begin{figure*}[t]
    \centering
    \includegraphics[width=.9\linewidth]{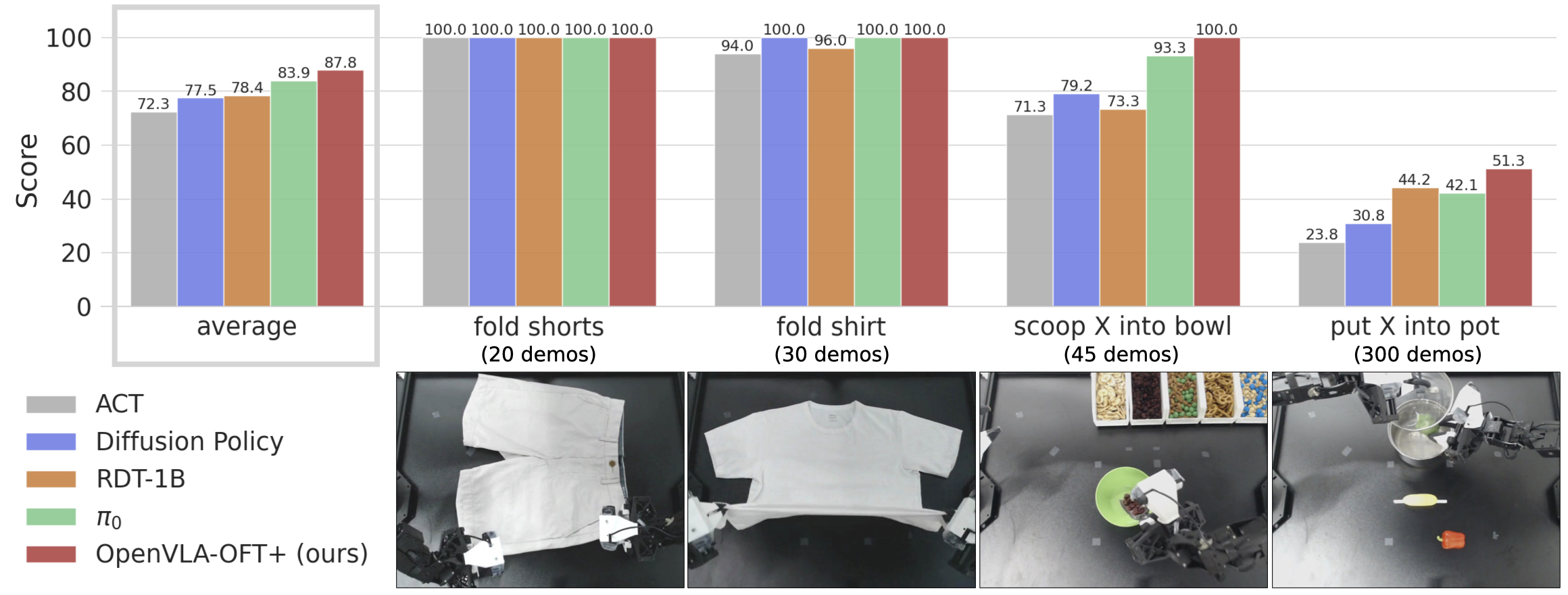}
    \caption{\textbf{ALOHA task performance results.} Comparison between policies trained from scratch (ACT, Diffusion Policy) and fine-tuned VLAs (RDT-1B, $\pi_0$, OpenVLA-OFT+) across four ALOHA manipulation tasks. OpenVLA-OFT+ enhances the base model with parallel decoding, action chunking, continuous actions, L1 regression, and FiLM \cite{perez2018film} for language grounding. Fine-tuned VLAs consistently outperform from-scratch methods, with OpenVLA-OFT+ achieving the highest average performance. Scores represent average percent completion of each task (see rubrics and detailed results in Appendix \ref{app:aloha_scoring_rubrics}).
    }
    \label{fig:aloha_performance_results}
\end{figure*}

\section{Experiments: Adapting OpenVLA to a Real-World ALOHA Robot}
\label{sec:expts_adapting_to_aloha}

While our experimental results in the prior section demonstrate OpenVLA-OFT's effectiveness in simulation, successful deployment in the real world, on robot platforms that differ substantially from those seen during pretraining, is crucial for showing broad applicability. We thus assess the efficacy of our optimized fine-tuning recipe on the ALOHA robot setup \citep{zhao2023learning}, a real bimanual manipulation platform operating at a high control frequency. We evaluate on novel dexterous manipulation tasks that have never been encountered before during OpenVLA's pretraining (which only involves single-arm robot data).

Prior works \citep{wen2024tinyvla, liu2024rdt, black2024pi_0} have shown that vanilla LoRA fine-tuning with autoregressive VLAs \citep{kim2024openvla} is impractical for such tasks, as its throughput (3-5 Hz for single-arm robots and even lower for bimanual tasks) falls well below the 25-50 Hz required for real-time deployment. We therefore exclude this baseline from our experiments and compare more effective methods that we discuss shortly.

In this section, we use an augmented version of our VLA fine-tuning recipe (\textbf{OFT+}) that additionally includes feature-wise linear modulation (FiLM) for enhanced language grounding, as described in Section \ref{sec:implementing_film}. We denote the OpenVLA policy instantiated through this augmented fine-tuning recipe as \textbf{OpenVLA-OFT+}.

\subsection{ALOHA Experimental Setup}

The ALOHA platform comprises two ViperX 300 S arms, three camera viewpoints (one top-down, two wrist-mounted), and robot state inputs (14-dimensional joint angles). It operates at 25 Hz (reduced from the original 50 Hz to enable faster training while still maintaining smooth robotic control), with actions representing target absolute joint angles. This setup differs significantly from OpenVLA's pretraining, which includes single-arm robot data only, a single camera viewpoint from a third-person camera, no robot state inputs, low-frequency control (3-10 Hz), and relative end-effector pose actions. The distribution shift poses a challenge to the adaptation of this model.

We design four representative tasks testing deformable object manipulation, long-horizon skills, tool usage, and language-driven control:
\begin{enumerate}
    \item \textbf{``fold shorts''}: Fold white shorts on a table with two consecutive bimanual folds. Training: 20 demonstrations. Evaluation: 10 trials.
    \item \textbf{``fold shirt''}: Fold white T-shirt through multiple synchronized bimanual folds, testing contact-rich, long-horizon manipulation. Training: 30 demonstrations. Evaluation: 10 trials.
    \item \textbf{``scoop X into bowl''}: Move bowl to center of table with left arm, scoop specified ingredient (``raisins,'' ``almonds and green M\&Ms,'' or ``pretzels'') with right arm using metal spoon. Training: 45 demonstrations (15 per ingredient). Evaluation: 12 trials (4 per ingredient).
    \item \textbf{``put X into pot''}: Open pot with left arm, place specified item (``green pepper,'' ``red pepper,'' or ``yellow corn'') with right arm, close pot. Training: 300 demonstrations (100 per object). Evaluation: 24 trials (12 in-distribution, 12 out-of-distribution).
\end{enumerate}

We fine-tune OpenVLA using OFT+ on each task independently for 50-150K gradient steps (total batch size 32 with 8 A100/H100-80GB GPUs) with action chunk size $K=25$. At inference time, we execute the full action chunk before requerying the model for the next chunk.

\subsection{Methods in Comparison}

The ALOHA tasks present a significant adaptation challenge for OpenVLA as the base model, given the substantial differences from its pretraining platforms in terms of control frequency, action space, and input modalities. For this reason, we compare OpenVLA-OFT+ against more recent VLAs---RDT-1B \citep{liu2024rdt} and $\pi_0$ \citep{black2024pi_0}---that were pretrained on bimanual manipulation data and might reasonably be expected to perform better on these downstream tasks. We evaluate these models after fine-tuning them using their authors' recommended recipes, and these methods serve as important points of comparison. Additionally, to provide comparisons with computationally efficient alternatives, we evaluate two popular imitation learning baselines: ACT \citep{zhao2023learning} and Diffusion Policy \citep{chi2023diffusion}, trained from scratch on each task.

To enable language following in these baseline methods, we use language-conditioned implementations. For ACT, we modify EfficientNet-B0 \citep{tan2019efficientnet} to process CLIP \citep{radford2021learning} language embeddings via FiLM \citep{perez2018film, shi2024yell}.\footnote{We use this FiLM-EfficientNet implementation only for language-dependent tasks (``scoop X into bowl'' and ``put X into pot''). For clothes folding tasks, we use the original ResNet-18 \citep{he2016deep} backbone as in \citep{zhao2024aloha}.} For Diffusion Policy, we use the DROID dataset \citep{khazatsky2024droid} implementation that conditions action denoising on DistilBERT \citep{sanh2019distilbert} language embeddings, modified to support bimanual control and multiple image inputs.

\subsection{ALOHA Task Performance Results}

\begin{figure}[t]
    \centering
    \includegraphics[width=\linewidth]{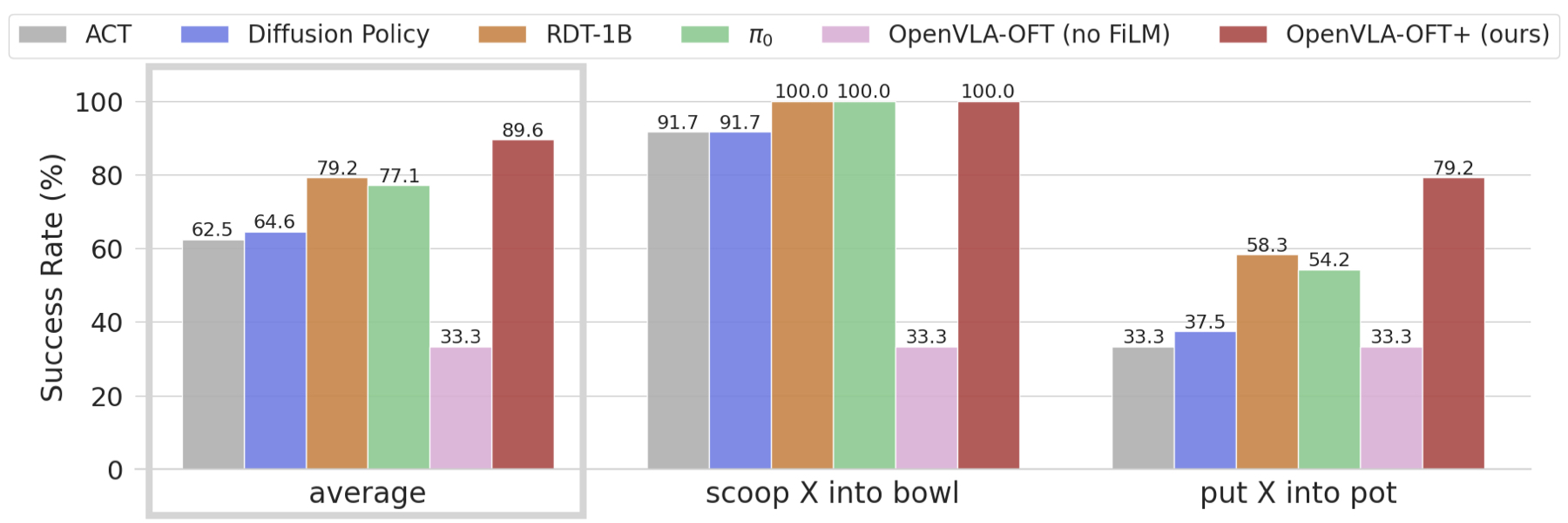}
    \caption{\textbf{ALOHA language following results.} Success rates in approaching language-specified target objects for language-dependent tasks. Fine-tuned VLAs follow the user's command more frequently than policies trained from scratch. OpenVLA-OFT+ shows the strongest language grounding, though removing FiLM \citep{perez2018film} reduces success to chance level.
    }
    \label{fig:aloha_lang_grounding_results}
\end{figure}

We evaluate all methods---ACT, Diffusion Policy, RDT-1B, $\pi_0$, and OpenVLA-OFT+---on our four ALOHA tasks. To provide fine-grained assessment, we use a predetermined rubric that assigns scores for partial task completion (see Appendix \ref{app:aloha_eval_details} for details). Figure \ref{fig:aloha_performance_results} shows aggregate performance scores, while Figure \ref{fig:aloha_lang_grounding_results} specifically tracks language following ability for the language-dependent tasks.

\textbf{Performance of non-VLA baselines.} The baseline methods trained from scratch show varying levels of success. ACT, while able to complete basic tasks, produces less precise actions and achieves the lowest overall performance. Diffusion Policy demonstrates stronger capabilities, matching or exceeding RDT-1B's reliability on the clothes folding and scooping tasks. However, it struggles with the ``put X into pot'' task which has a larger training dataset, suggesting limited scalability compared to VLA-based approaches.

\begin{figure}[t]
    \centering
    \includegraphics[width=\linewidth]{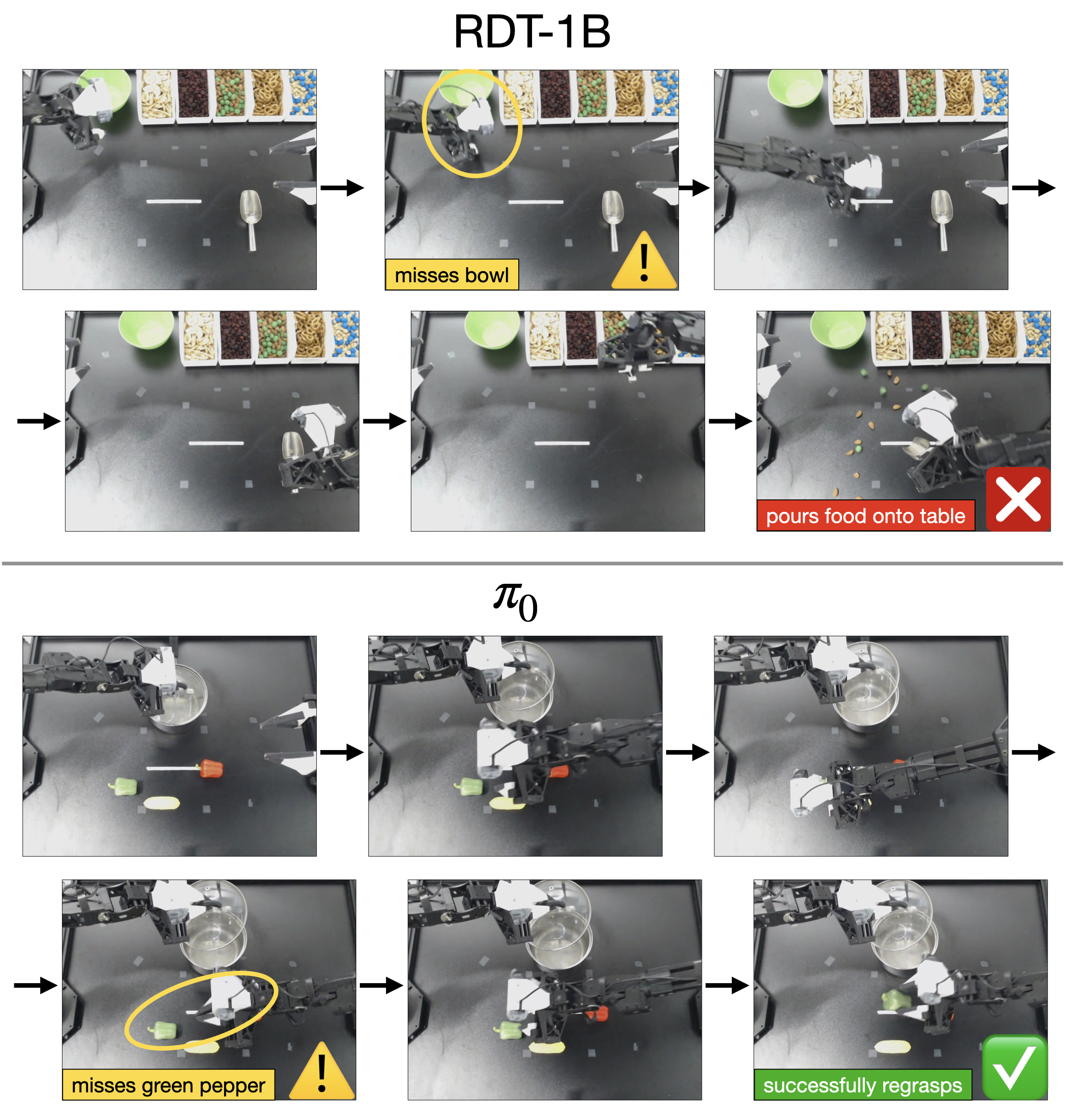}
    \caption{\textbf{Sample rollouts contrasting RDT-1B and $\pi_0$ error handling in ALOHA tasks.} \textbf{Top}: In some cases, RDT-1B fails to respond to missed bowl placement, continuing to pour ingredients into empty space. \textbf{Bottom}: $\pi_0$ demonstrates adaptive behavior by reattempting green pepper grasp after initial failure, highlighting better visual feedback integration. See \href{https://openvla-oft.github.io}{our website} for videos.
    }
    \label{fig:rdt_pi0_rollout}
\end{figure}

\begin{figure}[t]
    \centering
    \includegraphics[width=.95\linewidth]{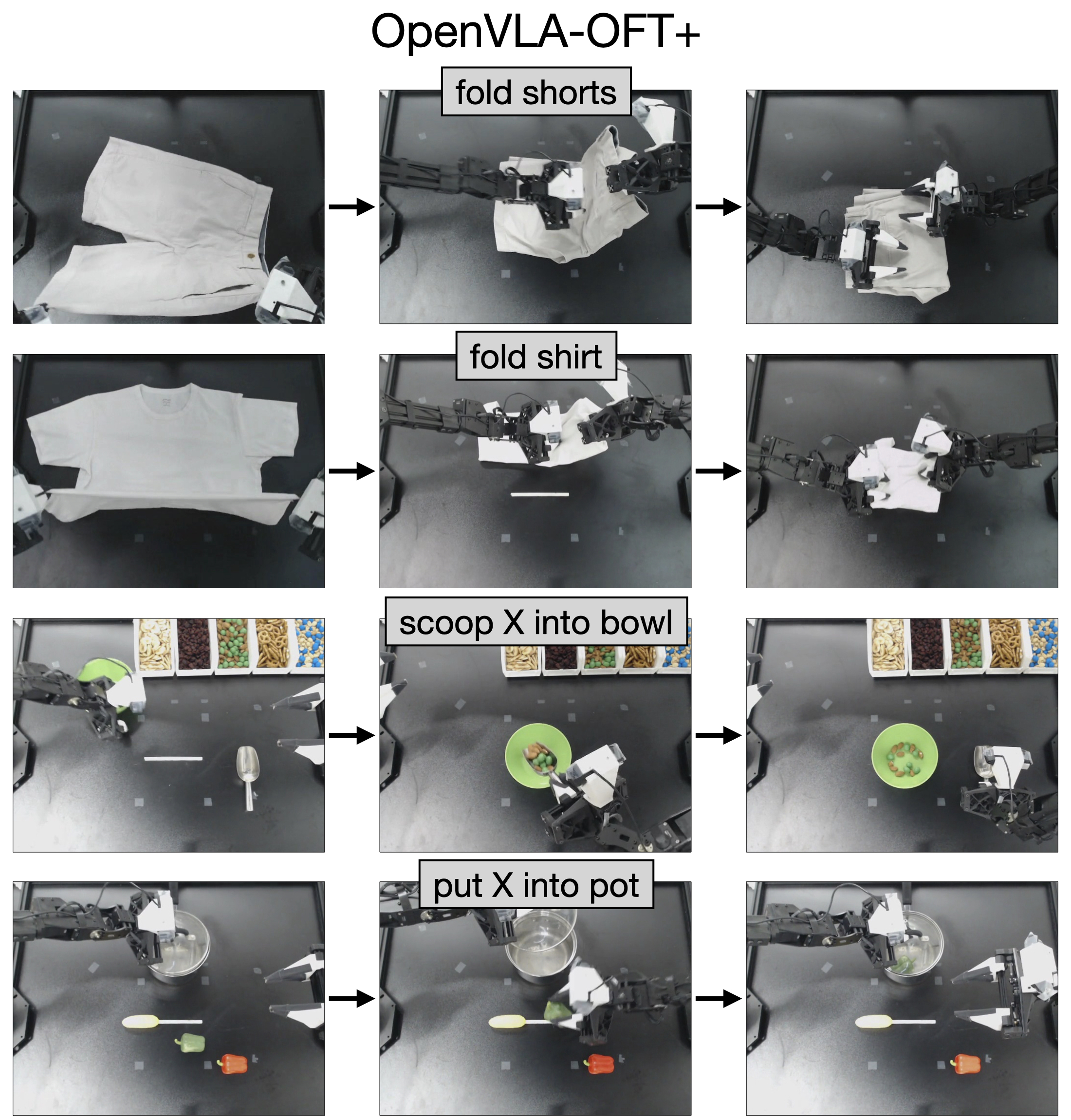}
    \caption{\textbf{Sample successful OpenVLA-OFT+ rollouts on the ALOHA robot.} OpenVLA-OFT+ can fold clothes, use a metal spoon to scoop and pour targeted trail mix ingredients into a bowl, and place targeted objects into a pot. See \href{https://openvla-oft.github.io}{our website} for videos.
    }
    \label{fig:openvla_rollouts}
\end{figure}

\textbf{Performance of fine-tuned VLAs.} Fine-tuned VLA policies generally outperform the from-scratch baselines in both task execution and language following, consistent with prior findings \citep{liu2024rdt, black2024pi_0}. Among VLAs, we observe distinct characteristics: RDT-1B achieves good language following through its ``Alternating Condition Injection'' scheme \citep{liu2024rdt}, but shows a limitation in handling closed-loop feedback. As visualized in Figure \ref{fig:rdt_pi0_rollout}, it often fails to correct mistakes in the ``scoop X into bowl'' task---for instance, continuing to pour ingredients into an imaginary bowl after missing the actual bowl, suggesting over-reliance on proprioceptive state over visual feedback. On the other hand, $\pi_0$ demonstrates more robust execution with smoother motions and better reactivity to feedback, often successfully recovering from initial failures (as shown in Figure \ref{fig:rdt_pi0_rollout}). While its language following slightly trails RDT-1B's, $\pi_0$ achieves better overall task completion, making it the strongest baseline. Finally, OpenVLA-OFT+ achieves the highest performance across both task execution and language following (see Figure \ref{fig:openvla_rollouts} for examples of successful task rollouts). This is particularly noteworthy given that the base OpenVLA model was pretrained only on single-arm data, while RDT-1B and $\pi_0$ were pretrained on substantial bimanual datasets (6K episodes and 8K hours of bimanual data, respectively). This suggests that the fine-tuning technique can be more crucial than pretraining data coverage for downstream performance.

\textbf{Ablation study of FiLM.} We evaluate the importance of FiLM in our OpenVLA-OFT+ approach by ablating it and assessing the policies' language following ability on the last two tasks, which require good language grounding for successful execution. As shown in Figure \ref{fig:aloha_lang_grounding_results}, language following drops to 33\% in both tasks---equal to randomly choosing the correct instruction. This demonstrates that FiLM is essential for preventing the model from overfitting to spurious visual features and ensuring proper attention to language inputs.

Please see the project website for ALOHA robot rollout videos and an in-depth qualitative analysis of all methods: \website

\begin{table}
\caption{\textbf{ALOHA robot inference efficiency results.} Throughput and latency measurements averaged over 100 queries on an NVIDIA A100 GPU. Each query processes three 224 × 224 px images, 14-D robot state, and a task command (``scoop raisins into bowl''). All methods use action chunk size K=25, except Diffusion Policy (K=24, as a multiple of 4 is required) and the original OpenVLA (K=1, no chunking). RDT-1B predicts 64 actions, but we execute the first 25 only for fair comparison. All implementations use PyTorch except $\pi_0$ (JAX). Bold and underlined values show best and second-best performance.}
\label{tab:aloha_efficiency_results}
\resizebox{0.5\textwidth}{!}{
\begin{tabular}{lcc}
\toprule
& Throughput (Hz) $\uparrow$ & Latency (Sec) $\downarrow$ \\
\midrule
OpenVLA & 1.8 & 0.543 \\
OpenVLA-OFT+ & 77.9 & 0.321 \\
RDT-1B & 84.1 & 0.297 \\
Diffusion Policy & 267.4 & 0.090 \\
$\pi_0$ & \underline{291.6} & \underline{0.086} \\
ACT & \textbf{432.8} & \textbf{0.058} \\
\bottomrule
\end{tabular}}
\end{table}

\subsection{ALOHA Inference Efficiency Comparisons}

We evaluate inference efficiency by measuring action generation throughput and latency across 100 queries for each method. We report the results in Table \ref{tab:aloha_efficiency_results}. The original OpenVLA formulation, even with just the additional wrist camera inputs, shows poor efficiency with 1.8 Hz throughput and 0.543 sec latency. In contrast, OpenVLA-OFT+ achieves 77.9 Hz throughput, though its latency is higher compared to the policies in the prior LIBERO experiments since it must process two additional input images.

Other methods demonstrate higher throughput than OpenVLA-OFT+ due to their smaller architectures: ACT (84M parameters), Diffusion Policy (157M), RDT-1B (1.2B), and $\pi_0$ (3.3B)---while OpenVLA has 7.5B parameters. ACT achieves the highest speed by combining L1 regression-based single-pass action generation (like OpenVLA-OFT+) with its compact architecture. Also, despite its larger size, $\pi_0$ outperforms both RDT-1B and Diffusion Policy in speed thanks to its optimized JAX implementation (all other methods are implemented in PyTorch).

Notably, OpenVLA-OFT+'s throughput (77.9 Hz) approaches RDT-1B's (84.1 Hz) despite being 7$\times$ larger, as it generates actions in a single forward pass rather than requiring multiple denoising steps as in RDT-1B.

\section{Discussion}
\label{sec:discussion}

Our study on VLA fine-tuning design decisions reveals how different components impact inference efficiency, task performance, model input-output flexibility, and language following ability. These insights lead to our Optimized Fine-Tuning (OFT) recipe, which enables effective VLA adaptation to novel robots and tasks through parallel decoding, action chunking, continuous actions, L1 regression, and (optionally) FiLM language conditioning. The success of OFT is particularly noteworthy with OpenVLA: despite having no exposure to bimanual robots or multi-view image inputs during pretraining, OpenVLA fine-tuned with OFT can adapt to such configurations and match or even outperform more recent diffusion-based VLAs ($\pi_0$ and RDT-1B) which \emph{have} encountered bimanual manipulators and multiple input images during pretraining. This demonstrates that a well-designed fine-tuning recipe can have a significant impact on final performance, and existing VLAs can be successfully adapted to new robotic systems without extensive retraining from scratch. Moreover, our results show that a simple L1 regression-based approach with a high-capacity model such as OpenVLA is quite effective for adapting to novel robots and tasks. This approach offers practical advantages over diffusion-based methods: the simpler algorithm leads to faster training convergence and inference speed while maintaining strong performance, making it particularly suitable for real-world robotics applications.

\section{Limitations}
\label{sec:Limitations}

While our Optimized Fine-Tuning (OFT) recipe shows promise for adapting VLAs to novel robots and tasks, several important questions remain.

\textbf{Handling multimodal demonstrations.} Our experiments use focused demonstration datasets with a consistent strategy per task. While L1 regression may help smoothen out noise in training demonstrations by encouraging the policy to learn the median mode in demonstrated actions, it may struggle to accurately model truly multimodal action distributions where multiple valid actions exist for the same input, which may not be ideal in cases where the ability to generate alternative action sequences would be beneficial for task completion. Conversely, diffusion-based approaches may better capture such multimodality but risk overfitting to suboptimal modes in training data (see \href{https://openvla-oft.github.io/#l1-regression-vs-diffusion}{our website} for discussions and video illustrations of these nuances). Understanding OFT's effectiveness with multimodal demonstrations remains an important direction for future work.

\textbf{Pretraining versus fine-tuning.} Our study focuses specifically on \emph{fine-tuning} VLAs for downstream tasks. Whether OFT's benefits extend effectively to \emph{pretraining}, or whether more expressive algorithms like diffusion are necessary for large-scale training, requires further investigation.

\textbf{Inconsistent language grounding.} Our ALOHA experiments reveal that OpenVLA without FiLM exhibits poor language grounding, despite showing no such issues in LIBERO simulation benchmark experiments. The source of this discrepancy---whether from the lack of bimanual data in pretraining or other factors---remains unclear and warrants further study.

\section*{Acknowledgments}

We thank the Stanford Center for Research on Foundation Models (CRFM) and Stanford Institute for Human-Centered AI (HAI) for providing computational resources that supported this research. Toyota Research Institute (TRI) provided funds to assist the authors with their research, but this article solely reflects the opinions and conclusions of its authors and not TRI or any other Toyota entity. This work was also in part supported by the Robotics and AI Institute and ONR grant N00014-22-1-2621. We thank Physical Intelligence for providing beta access to the $\pi_0$ model used in our ALOHA robot evaluations. We also thank Dan Fu, Karl Pertsch, and Alec Lessing for insightful discussions which contributed to the development of this work. Lastly, we are grateful to Moritz Reuss for providing additional experimental results for MDT and engaging in helpful discussions related to this work.

\bibliographystyle{plainnat}
\bibliography{references}

\clearpage
\appendix
\section{Appendix}

\begin{figure*}[t]
    \centering
    \includegraphics[width=\linewidth]{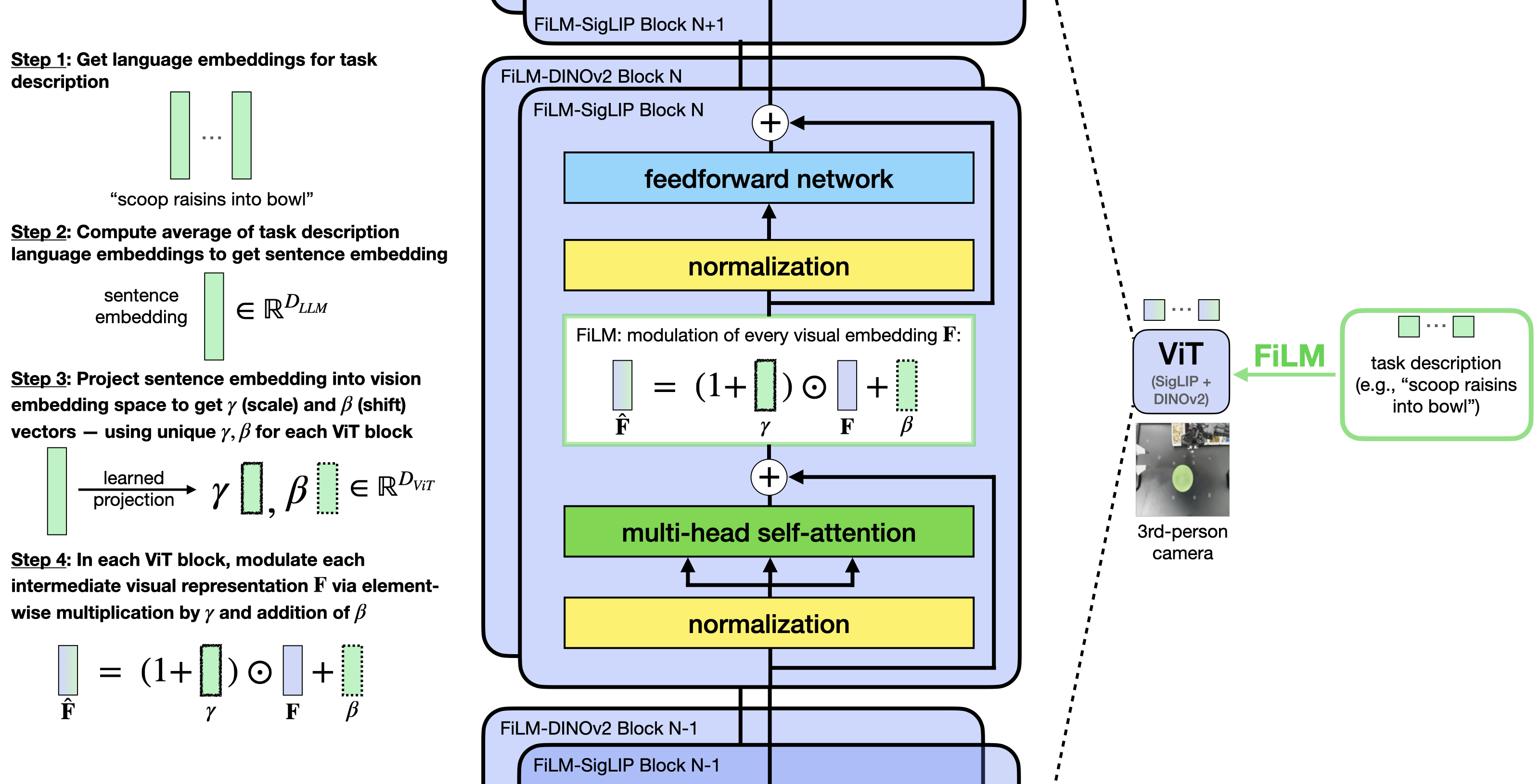}
    \caption{\textbf{Feature-wise linear modulation (FiLM) implementation in OpenVLA's vision backbone.} We integrate FiLM \citep{perez2018film} into both SigLIP \citep{zhai2023sigmoid} and DINOv2 \citep{oquab2023dinov2} vision transformers in OpenVLA's fused vision backbone. The average task description embedding modulates visual features through scale and shift operations at each transformer block, enhancing language-vision integration. This modification significantly improves language following in ALOHA tasks.}
    \label{fig:film}
\end{figure*}

\subsection{Model Architecture Details}
\label{app:model_architecture_details}

\textbf{Base OpenVLA Architecture.} OpenVLA combines a fused vision backbone (with both SigLIP \citep{zhai2023sigmoid} and DINOv2 \citep{oquab2023dinov2} vision transformers), a Llama-2 7B language model \citep{touvron2023llama}, and a 3-layer MLP projector with GELU activation \cite{hendrycks2016gaussian} for projecting visual features into the language embedding space.

The original model processes a single third-person image and a language instruction (e.g., ``put eggplant into pot''). The fused vision encoder extracts 256 patch embeddings from each vision transformer, concatenates them along the hidden dimension, and projects them into the language embedding space. These projected features are concatenated with language embeddings along the sequence dimension before being processed by the Llama-2 decoder to output a 7-dimensional robot action representing delta end-effector pose, represented by a string of discrete action tokens.

\textbf{OpenVLA-OFT architecture modifications.} OpenVLA-OFT introduces six key changes:
\begin{enumerate}
    \item processes multiple input images (e.g., third-person image plus wrist camera images) through the shared SigLIP-DINOv2 backbone
    \item projects robot proprioceptive state to language embedding space via a 2-layer MLP with GELU activation
    \item replaces causal attention with bidirectional attention for parallel decoding
    \item substitutes the language model decoder output layer with a 4-layer MLP (ReLU activation) for generation of continuous actions (instead of discrete actions)
    \item outputs chunks of K actions instead of single-timestep actions
    \item (for OpenVLA-OFT+) adds FiLM \citep{perez2018film} modules that use the average task language embedding to modulate visual features in both SigLIP and DINOv2 vision transformers (see Appendix \ref{app:film_details} for details)
\end{enumerate}

The complete OpenVLA-OFT+ architecture is illustrated in Figure \ref{fig:openvla_aloha}.

\subsection{Implementation Details}
\label{app:implementation_details}

\subsubsection{Parallel Decoding Implementation}
In the original OpenVLA autoregressive training scheme, the model receives ground-truth action tokens shifted right by one position as input (a setup known as teacher forcing). A causal attention mask ensures the model only attends to current and previous tokens. At test time, each predicted token is fed back as input for the next prediction.

For parallel decoding, we replace this input with empty action embeddings that differ only in their positional encoding values (similar to \citep{zhao2023learning}). We also use a bidirectional attention mask (instead of causal), enabling the model to leverage all intermediate features non-causally when predicting each element in the action chunk.

\subsubsection{Continuous Action Representations}
For discrete actions, increasing the number of bins used for discretization improves precision but reduces the frequency of individual tokens in the training data, potentially hurting generalization. On the other hand, with a continuous action representation, the VLA can directly model the action distribution without lossy discretization.

Our continuous representation implementations use the following specifications:

\textbf{L1 regression:} The MLP action head consists of 4 layers with ReLU activation, mapping final Llama-2 decoder layer hidden states directly to continuous actions.

\textbf{Diffusion:} We use:
\begin{itemize}
    \item DDIM \citep{song2020denoising} sampler with 50 diffusion timesteps
    \item Squared cosine beta schedule following \citep{chi2023diffusion, zhao2024aloha}
    \item 4-layer noise predictor with same MLP architecture as the L1 regression head
\end{itemize}

\subsubsection{Input Processing Details}
Passing each input image through the OpenVLA fused vision encoder produces 256 patch embeddings, which are projected to the langauge model embedding space via a 3-layer MLP with GELU activation \citep{hendrycks2016gaussian}. Low-dimensional robot states are also projected to the language embedding space through a 2-layer MLP with GELU activation.

\subsection{Feature-wise Linear Modulation (FiLM) Implementation Details}
\label{app:film_details}

\textbf{FiLM schematic.} Section \ref{sec:implementing_film} describes how we implement feature-wise linear modulation (FiLM) \citep{perez2018film} for OpenVLA. Figure \ref{fig:film} illustrates our implementation. OpenVLA has a fused vision encoder with both SigLIP \citep{zhai2023sigmoid} and DINOv2 \citep{oquab2023dinov2} vision transformers, and we apply FiLM to both transformers.

\textbf{Design considerations.} In our implementation, following \citet{perez2018film}, we multiply $\mathbf{F}$ by $(1+\mathbf{\gamma})$ instead of $\mathbf{\gamma}$ since $\gamma$ and $\beta$ are near zero at initialization. This helps preserve the visual encoder's original activations at the start of fine-tuning, minimizing perturbation in the pretrained representation.

\textbf{Implementation specifics.} The functions $f(\mathbf{x})$ and $h(\mathbf{x})$ that project language embeddings to obtain $\mathbf{\gamma}$ and $\mathbf{\beta}$ are implemented as simple affine transformations. Separate projectors are learned for each transformer block to allow for block-specific modulation patterns. This design enables the model to learn different modulation patterns at different levels of visual feature processing.

One might initially consider modulating each patch embedding independently, as opposed to each hidden dimension of each embedding as discussed in Section \ref{sec:implementing_film}. However, our spatially-agnostic modulation approach more closely mirrors FiLM's operation in convolutional networks, where modulation applies globally across spatial dimensions since entire feature maps are scaled and shifted by individual elements of $\mathbf{\gamma}$ and $\mathbf{\beta}$. This design choice better maintains the benefits of FiLM and improves the policy's language grounding substantially. We find that an alternative formulation that modulates each patch embedding independently leads to weaker language grounding.

\subsection{OpenVLA-OFT Hyperparameters and Training Details}
\label{app:openvla_oft_training_details}

\textbf{OpenVLA-OFT training details for LIBERO.} Hyperparameters for OpenVLA-OFT fine-tuning on LIBERO are listed in Table \ref{tab:openvla_oft_hyperparameters_libero}. We train until the mean L1 loss between predicted and ground-truth normalized actions (scaled between $[-1,+1]$) falls below 0.01. For faster convergence, we decay the learning rate from 5e-4 to 5e-5 after 100K gradient steps. We evaluate checkpoints every 50K steps, with the 150K checkpoint achieving best performance in all task suites except for LIBERO-Goal. Note that we do not use FiLM for LIBERO experiments since the fine-tuned policies without it already demonstrate good language grounding.

\textbf{OpenVLA-OFT+ training details for ALOHA.} Hyperparameters for OpenVLA-OFT+ training on ALOHA tasks (with FiLM in the augmented OFT+ recipe) are shown in Table \ref{tab:openvla_oft_hyperparameters_aloha}. We maintain the same convergence criterion as in the LIBERO experiments (training until mean normalized L1 loss falls below 0.01) and similar learning rate decay strategy (again 10$\times$ reduction, but after 50K gradient steps instead of 100K, since the ALOHA datasets are smaller). For the two clothes folding tasks which do not require language grounding, we still include FiLM to verify that these additional parameters do not impair task execution.

\subsection{Baseline Methods Hyperparameters and Training Details}
\label{app:baseline_training_details}

\textbf{ACT training details for ALOHA.} Table \ref{tab:act_hyperparameters_aloha} lists hyperparameters for ACT \citep{zhao2023learning} trained from scratch on each task. For non-language-dependent tasks (``fold shorts'', ``fold shirt''), we use the default ResNet-18 \citep{he2016deep} backbone, which does not include language conditioning. For language-dependent tasks (``scoop X into bowl'', ``put X into pot''), we implement EfficientNet-B0 \citep{tan2019efficientnet} with FiLM \citep{perez2018film}, similar to \citep{shi2024yell}. While the authors of ACT recommend training for at least 5K epochs, we extend training to 10K-70K epochs per task to improve performance.

\textbf{Diffusion Policy training details for ALOHA.} For Diffusion Policy training, we use the DROID implementation \citep{khazatsky2024droid}, which conditions action predictions on DistilBERT \citep{sanh2019distilbert} language embeddings of the task description. We list hyperparameters in Table \ref{tab:diffusion_policy_hyperparameters_aloha}.

\textbf{RDT-1B training details for ALOHA.} Hyperparameters for RDT-1B fine-tuning are shown in Table \ref{tab:rdt_hyperparameters_aloha}. The authors of RDT-1B recommend training for 150K gradient steps, but we observe that training converges in significantly fewer steps since our fine-tuning datasets are much smaller than the RDT-1B fine-tuning dataset. Therefore, we observe that it is unnecessary to train for such a large amount of time. In fact, on ``scoop X into bowl'', the earlier 18K step checkpoint (73.3\% success) outperforms the later 40K step checkpoint (70.0\%) (we report the former in our ALOHA experiments).

\textbf{$\pi_0$ training details for ALOHA.} Table \ref{tab:pi0_hyperparameters_aloha} lists hyperparameters for $\pi_0$ fine-tuning. We use full fine-tuning (the default option in their codebase) and train until convergence.

\subsection{ALOHA Evaluation Details}
\label{app:aloha_eval_details}

\subsubsection{ALOHA Task Suite Details}

\begin{figure*}[h]
    \centering
    \includegraphics[width=\linewidth]{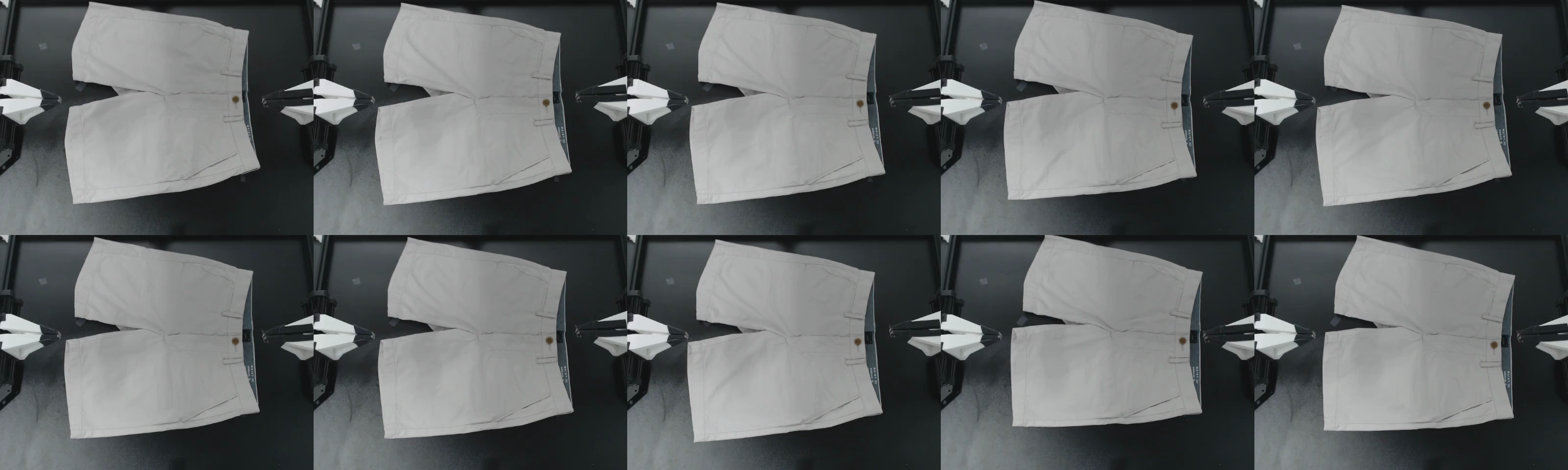}
    \caption{\textbf{Initial states for ``fold shorts'' task evaluations.} We use the smallest variance in initial positions for this task due to the small number of training demonstrations (20) compared to other tasks. During the first eight rollouts, the shorts' position is incrementally adjusted by up to 3 cm along the vertical axis of the tabletop surface. In the final two rollouts, the shorts are rotated 10 degrees clockwise to align the bottom edge to be parallel to the tabletop's horizontal axis.}
    \label{fig:fold_shorts_initial_positions}
\end{figure*}

Below are detailed specifications for each task in our ALOHA experiments:

\textbf{1. ``fold shorts''}
\begin{itemize}
    \item Task: Bimanual folding of white shorts with two synchronized folds
    \item Dataset: 20 demonstrations (19 training, 1 validation)
    \item Episode length: 1000 timesteps (40 seconds)
    \item Evaluation: 10 trials
    \item Initial states: See Figure \ref{fig:fold_shorts_initial_positions}
\end{itemize}

\begin{figure*}[h]
    \centering
    \includegraphics[width=\linewidth]{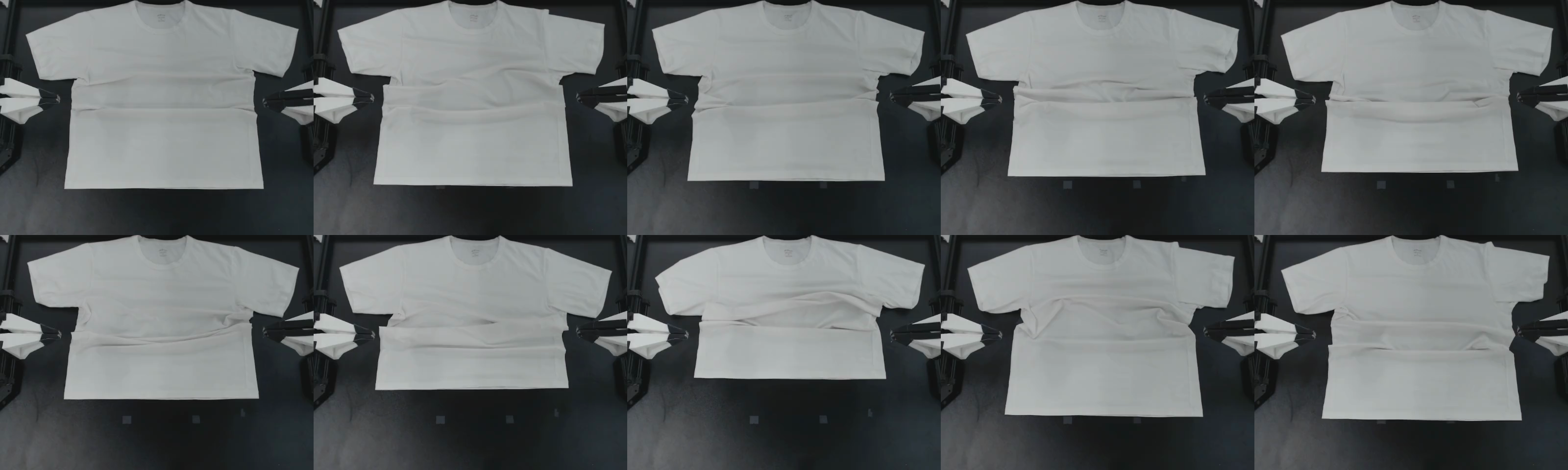}
    \caption{\textbf{Initial states for ``fold shirt'' task evaluations.} The bottom edge of the shirt varies by 15 cm along the tabletop's vertical axis across different trials.}
    \label{fig:fold_shirt_initial_positions}
\end{figure*}

\textbf{2. ``fold shirt''}
\begin{itemize}
    \item Task: Long-horizon T-shirt folding with multiple synchronized bimanual folds
    \item Dataset: 30 demonstrations (29 training, 1 validation)
    \item Episode length: 1250 timesteps (50 seconds)
    \item Evaluation: 10 trials
    \item Initial states: See Figure \ref{fig:fold_shirt_initial_positions}
\end{itemize}

\begin{figure*}[h]
    \centering
    \includegraphics[width=0.8\linewidth]{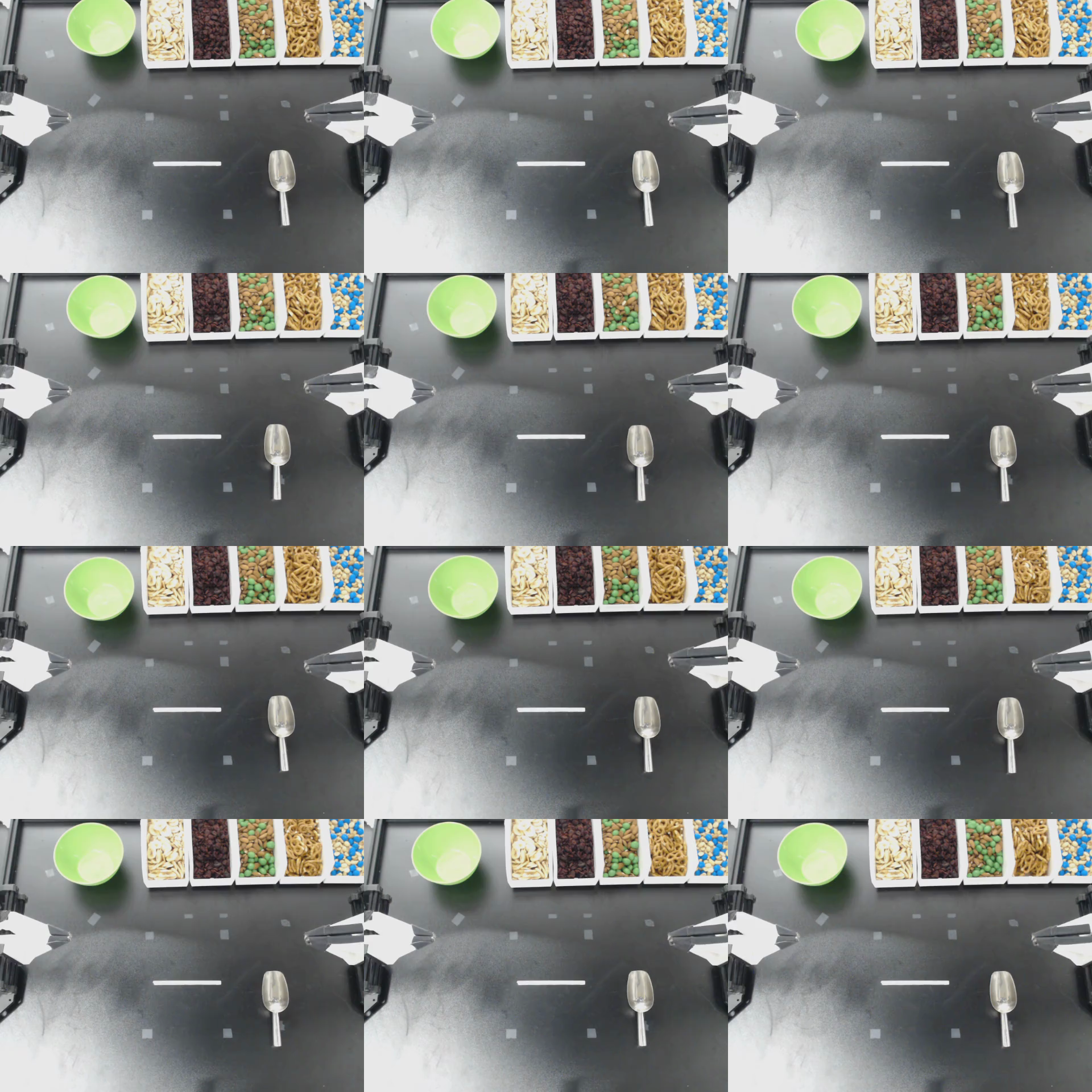}
    \caption{\textbf{Initial states for ``scoop X into bowl'' task evaluations.} The green bowl's position varies by up to 10 cm across both vertical and horizontal table axes. Simultaneously, the metal spoon's position varies by 3 cm along the horizontal axis. Each row in this grid displays a set of three identical initial states since we rotate among three target food ingredients (``raisins'', ``almonds and green M\&Ms'', and ``pretzels'') with the same initial states, before moving on to the next set of initial states.}
    \label{fig:scoop_x_into_bowl_initial_positions}
\end{figure*}

\textbf{3. ``scoop X into bowl''}
\begin{itemize}
    \item Task: Move bowl to center, scoop specified ingredient (raisins, almonds and green M\&Ms, or pretzels) into bowl
    \item Dataset: 45 demonstrations (15 per target; 42 training, 3 validation)
    \item Episode length: 900 timesteps (36 seconds)
    \item Evaluation: 12 trials
    \item Initial states: See Figure \ref{fig:scoop_x_into_bowl_initial_positions}
\end{itemize}

\begin{figure*}[h]
    \centering
    \includegraphics[width=0.8\linewidth]{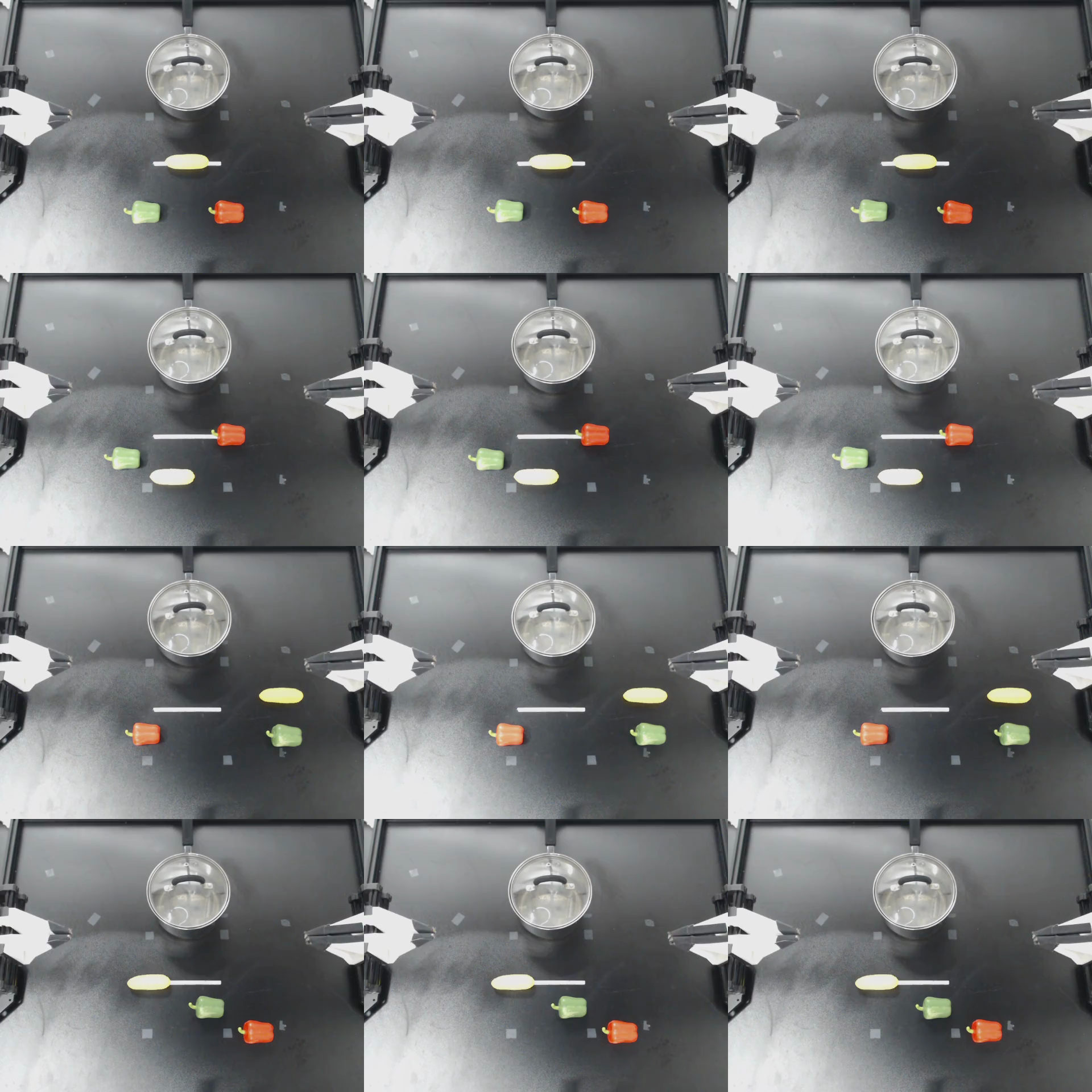}
    \caption{\textbf{Initial states for ``put X into pot'' task evaluations (in-distribution version).} Food object positions vary by 15 cm vertically and 40 cm horizontally across trials, while the pot remains stationary. Each grid row presents three identical initial states, as we cycle through the three target objects (``green pepper'', ``red pepper'', and ``yellow corn'') using the same initial configurations before we move on to the next set of initial configurations. (Refer to Figure \ref{fig:put_x_into_pot_ood_initial_positions} for the out-of-distribution version of this task.)}
    \label{fig:put_x_into_pot_initial_positions}
\end{figure*}

\begin{figure*}[h]
    \centering
    \includegraphics[width=0.8\linewidth]{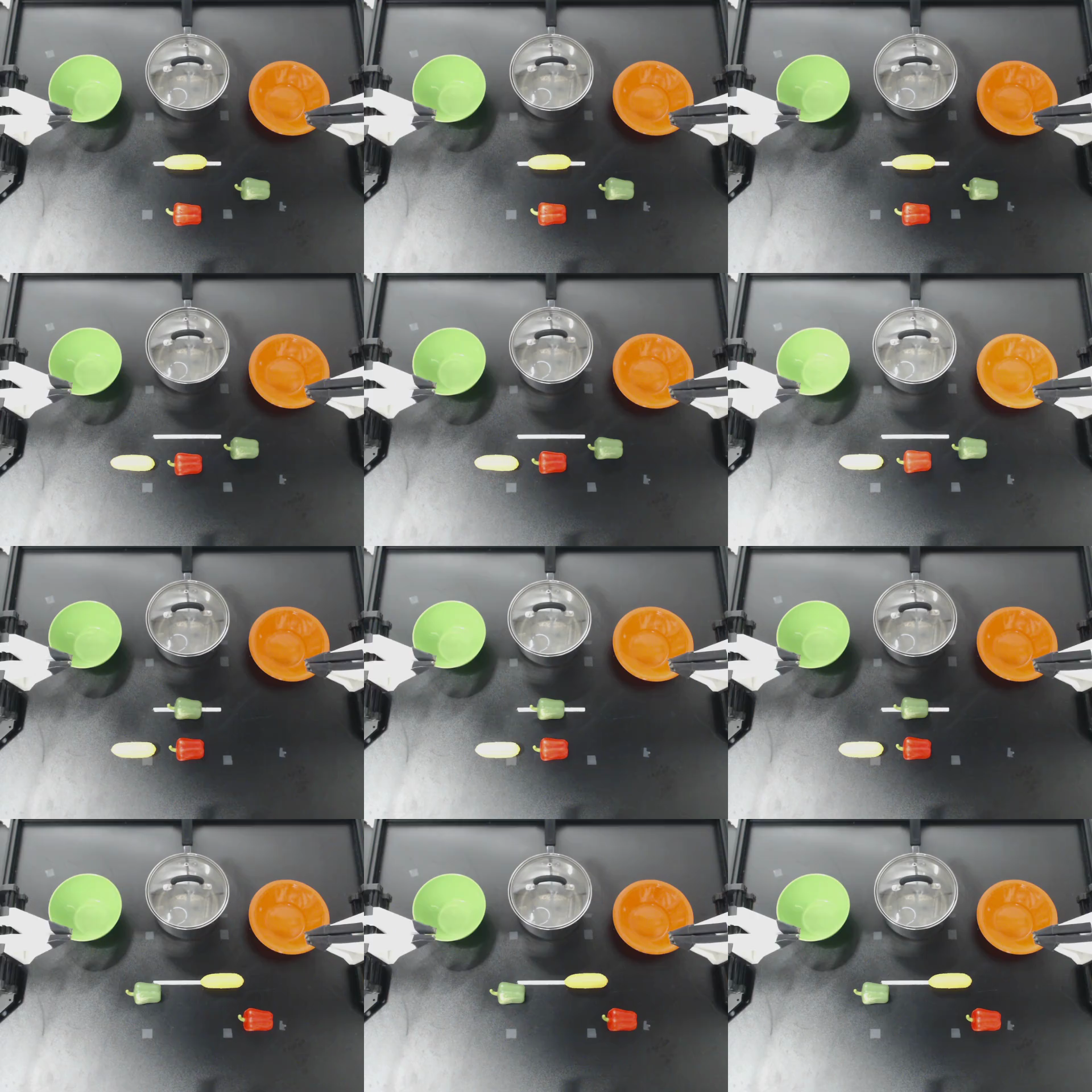}
    \caption{\textbf{Initial states for ``put X into pot'' task evaluations (out-of-distribution version).} Unlike the in-distribution version of this task (Figure \ref{fig:put_x_into_pot_initial_positions}), this version introduces unseen green and orange bowls as distractor objects to assess robustness to novel visual inputs. Food object positions vary by 15 cm vertically and 30 cm horizontally across trials, with the pot remaining stationary. Each grid row presents three identical initial states, as we cycle through the three target objects (``green pepper'', ``red pepper'', and ``yellow corn'') using the same initial configurations before we move on to the next set of initial configurations.}
    \label{fig:put_x_into_pot_ood_initial_positions}
\end{figure*}

\textbf{4. ``put X into pot''}
\begin{itemize}
    \item Task: Open pot, place specified item (green pepper, red pepper, or yellow corn) into pot, close pot
    \item Dataset: 300 demonstrations (100 per target; 285 training, 15 validation)\footnote{This relatively large number of demonstrations for the ``put X into pot'' task is not necessary for satisfactory performance. It simply reflects an earlier investigative phase of this work during which we encountered difficulties with language grounding in learned policies and initially hypothesized that increasing demonstration quantity might improve language following capabilities. However, merely expanding the training set proved insufficient for achieving satisfactory language grounding, and we discovered that additional techniques were needed for more reliable language grounding. We still decided to fine-tune on the full dataset with 300 demonstrations nonetheless.}
    \item Initial variation: 45 cm horizontal, 20 cm vertical for food items; fixed pot pose
    \item Episode length: 400 timesteps (16 seconds)
    \item Evaluation: 24 trials (12 in-distribution evaluations, 12 out-of-distribution evaluations)
    \item Initial states: See Figures \ref{fig:put_x_into_pot_initial_positions} (in-distribution) and \ref{fig:put_x_into_pot_ood_initial_positions} (out-of-distribution)
\end{itemize}

\subsubsection{ALOHA Task Scoring Rubric}
\label{app:aloha_scoring_rubrics}

The scoring rubrics and detailed results for the four ALOHA tasks are shown in Tables \ref{tab:detailed_aloha_fold_shorts_results}, \ref{tab:detailed_aloha_fold_shirt_results}, \ref{tab:detailed_aloha_scoop_bowl_results}, and \ref{tab:detailed_aloha_put_x_into_pot_results}.

\vspace{-0.1cm}

\subsection{Additional Experiments}
\label{app:additional_experiments}

\subsubsection{Single OpenVLA-OFT Policy for All LIBERO Task Suites Combined}
\label{app:single_policy_all_libero}

In Section \ref{sec:expts_evaluating_design_decisions} and Table \ref{tab:policy_performance_results}, we report results with OpenVLA-OFT policies trained on each task suite independently. In this section, we assess whether our method scales to larger fine-tuning datasets by training one OpenVLA-OFT policy on all four task suites combined. As shown in Table \ref{tab:additional_libero_results}, this new policy achieves comparable average task performance as the task suite-specific policies---confirming that our method scales to larger fine-tuning datasets.

\subsubsection{Ablating FiLM in LIBERO}
\label{app:ablating_film_libero}

The FiLM ablation study in Section \ref{sec:expts_adapting_to_aloha} suggests that FiLM is crucial for enabling strong language following in real-world ALOHA robot tasks. In this section, we assess whether FiLM is similarly important in the LIBERO simulation tasks. We train a single OpenVLA-OFT+ policy (with FiLM) on all LIBERO task suites combined (similar to the previous section) and report task performance results in Table \ref{tab:additional_libero_results}, comparing performance against the OpenVLA-OFT policy trained without FiLM. In LIBERO, FiLM leads to slightly higher average success rate, though the difference here is minor compared to the findings in the real-world ALOHA experiments, where language following is more challenging due to the reasons discussed in Section \ref{sec:implementing_film}.

\subsubsection{Ablating the OpenVLA Pretrained Representation}

We evaluate the performance of OpenVLA-OFT policies produced by fine-tuning the underlying Prismatic VLM \citep{karamcheti2024prismatic} directly on the LIBERO downstream datasets without OpenVLA's Open X-Embodiment \citep{o2023open} robot pretraining. This ablation study investigates whether OpenVLA's robot-pretrained representation remains valuable when subjected to a substantially different fine-tuning approach such as OFT. The results in Table \ref{tab:app:openvla_scratch_ablation_libero} demonstrate that the variant without the pretrained OpenVLA representation consistently underperforms compared to the full OpenVLA-OFT model, confirming the benefits of using the pretrained representation for downstream policy learning.

\subsubsection{Scaling Up OpenVLA-OFT to a Larger Real-World Dataset (BridgeData V2)}

In Appendix \ref{app:single_policy_all_libero}, we observe that a single OpenVLA-OFT policy can effectively fit all four LIBERO task suite datasets combined, confirming that the proposed method scales to larger fine-tuning datasets. In this section, we scale up the fine-tuning data further and assess whether OpenVLA-OFT can also fit a real-world robotic manipulation dataset that is significantly larger and more diverse than both the LIBERO datasets and the ALOHA robot datasets discussed in Section \ref{sec:expts_adapting_to_aloha}. Specifically, we train OpenVLA-OFT (without FiLM) on the BridgeData V2 dataset \citep{walke2023bridgedata}, which contains 50,365 real WidowX robot demonstrations (25$\times$ more than the four LIBERO task suites combined).\footnote{Although the base OpenVLA model was already pretrained on Bridge data, we note that the model must still be fine-tuned when using the new OFT recipe since the architecture, learning algorithm, and action representation and decoding scheme have been modified significantly from the pretraining setup. In fact, the initial action regression L1 loss in the beginning of OpenVLA-OFT training on Bridge is large (roughly 0.5 when actions are normalized to the scale $[-1,+1]$).}

We evaluate OpenVLA-OFT on a subset of BridgeData V2 WidowX robot tasks from the evaluation suite used in the original OpenVLA work \citep{kim2024openvla}. This representative subset covers the four types of generalization (visual, motion, physical, semantic) as well as language grounding tasks. We compare the task performance of OpenVLA-OFT with that of the public OpenVLA checkpoint, scoring both methods using the same criteria used in the OpenVLA work. As shown in Table \ref{tab:bridge_results}, OpenVLA-OFT surpasses OpenVLA on average across these tasks (69.2\% versus 65.8\%, respectively)---confirming scalability of the proposed OFT approach to much larger and more diverse real robot datasets. Further, we observe that FiLM is not necessary for satisfactory language following in Bridge tasks, likely because the base model has shown effective language following in Bridge tasks.

\begin{table*}
\centering
\caption{\textbf{OpenVLA-OFT hyperparameters for LIBERO.} We specify the hyperparameters for the full OpenVLA-OFT approach which includes parallel decoding, action chunking, continuous actions with L1 regression, and additional inputs (wrist camera and robot state).}
\label{tab:openvla_oft_hyperparameters_libero}
\begin{tabular}{ll}
\toprule
hyperparameter & value \\
\midrule
\# GPUs & 8 x NVIDIA A100 or H100 (80GB VRAM) \\
learning rate (LR) & 5e-4 \\
total batch size & 64 (8 per GPU) \\
\# train steps & 150K for LIBERO-Spatial (with LR=5e-5 after 100K steps); \\
& 150K for LIBERO-Object (with LR=5e-5 after 100K steps); \\
& 50K for LIBERO-Goal; \\
& 150K for LIBERO-Long (with LR=5e-5 after 100K steps) \\
input images & 1 third-person camera image, 1 wrist-mounted camera image \\
input image size & 224 x 224 px \\
use observation history & no (use single-step inputs) \\
LoRA rank & 32 \\
action chunk size & 8 steps (predict 8, execute all 8 open-loop at test time) \\
use proprio (robot state) & yes \\
use FiLM & no \\
\# trainable parameters & 279M total (111M LoRA adapter + 151M action head + 17M proprio projector) \\
image augmentations & 90\% random crops, color jitter: \\
& \;\; \texttt{random\_resized\_crop=dict(scale=[0.9, 0.9], ratio=[1.0, 1.0])} \\
& \;\; \texttt{random\_brightness=[0.2]} \\
& \;\; \texttt{random\_contrast=[0.8, 1.2]} \\
& \;\; \texttt{random\_saturation=[0.8, 1.2]} \\
& \;\; \texttt{random\_hue=[0.05]} \\
\bottomrule
\end{tabular}
\end{table*}

\begin{table*}
\centering
\caption{\textbf{OpenVLA-OFT+ hyperparameters for ALOHA experiments.} We specify the hyperparameters for the full OpenVLA-OFT+ approach which includes parallel decoding, action chunking, continuous actions with L1 regression, additional inputs (two wrist camera images and robot state), and FiLM.}
\label{tab:openvla_oft_hyperparameters_aloha}
\begin{tabular}{ll}
\toprule
hyperparameter & value \\
\midrule
\# GPUs & 8 x NVIDIA A100 or H100 (80GB VRAM) \\
learning rate (LR) & 5e-4 \\
total batch size & 32 (4 per GPU) \\
\# train steps & 100K for ``fold shorts'' (with LR=5e-5 after 50K steps); \\
& 70K for ``fold shirt'' (with LR=5e-5 after 50K steps); \\
& 50K for ``scoop X into bowl''; \\
& 100K for ``put X into pot'' (with LR=5e-5 after 50K steps); \\
input images & 1 third-person camera image, 2 wrist-mounted camera images (left wrist + right wrist) \\
input image size & 224 x 224 px \\
use observation history & no (use single-step inputs) \\
LoRA rank & 32 \\
action chunk size & 25 steps (predict 25, execute all 25 open-loop at test time) \\
use proprio (robot state) & yes \\
use FiLM & yes \\
\# trainable parameters & 853M total (111M LoRA adapter + 269M action head + 17M proprio projector + 456M FiLM projectors) \\
image augmentations & 90\% random crops, color jitter: \\
& \;\; \texttt{random\_resized\_crop=dict(scale=[0.9, 0.9], ratio=[1.0, 1.0])} \\
& \;\; \texttt{random\_brightness=[0.2]} \\
& \;\; \texttt{random\_contrast=[0.8, 1.2]} \\
& \;\; \texttt{random\_saturation=[0.8, 1.2]} \\
& \;\; \texttt{random\_hue=[0.05]} \\
\bottomrule
\end{tabular}
\end{table*}

\begin{table*}
\centering
\caption{\textbf{ACT hyperparameters for ALOHA experiments.} We follow default parameters from \citet{zhao2023learning} with three modifications: increased batch size (8 to 64), reduced chunk size (100 to 25 to match other methods in our experiments), and EfficientNet-B0 with FiLM \citep{perez2018film} for language-dependent tasks instead of ResNet-18. All models are trained from scratch for at least 5000 epochs, as recommended by the authors of ACT.}
\label{tab:act_hyperparameters_aloha}
\begin{tabular}{ll}
\toprule
hyperparameter & value \\
\midrule
\# GPUs & 1 x NVIDIA Titan RTX (24GB VRAM) \\
learning rate (LR) & 1e-5 \\
total batch size & 64 \\
\# train ``epochs'' & 70K for ``fold shorts'' \\
& 30K for ``fold shirt'' \\
& 20K for ``scoop X into bowl'' \\
& 10K for ``put X into pot'' \\
input images & 1 third-person camera image, 2 wrist-mounted camera images (left wrist + right wrist) \\
input image size & 224 x 224 px \\
use observation history & no (use single-step inputs) \\
action chunk size & 25 steps (predict 25, execute all 25 open-loop at test time) \\
use proprio (robot state) & yes \\
image backbone & ResNet-18 for ``fold shorts'' and ``fold shirt'' tasks; \\
& EfficientNet-B0 (w/ FiLM) for ``scoop X into bowl'' and ``put X into pot'' tasks \\
\# trainable parameters & 84M for ResNet-18 variant; 80M for EfficientNet-B0 w/ FiLM variant \\
\bottomrule
\end{tabular}
\end{table*}

\begin{table*}
\centering
\caption{\textbf{Diffusion Policy hyperparameters for ALOHA experiments.} This configuration follows the DROID implementation \citep{khazatsky2024droid} with two modifications: multiple input images for the ALOHA robot setup and larger image size (224 x 224 px) for fair comparison with other methods.}
\label{tab:diffusion_policy_hyperparameters_aloha}
\begin{tabular}{ll}
\toprule
hyperparameter & value \\
\midrule
\# GPUs & 1 x NVIDIA L40S (48GB VRAM) \\
learning rate (LR) & 1e-4 \\
total batch size & 128 \\
\# train steps & 30K for ``fold shorts'' \\
& 120K for ``fold shirt'' \\
& 80K for ``scoop X into bowl'' \\
& 40K for ``put X into pot'' \\
input images & 1 third-person camera image, 2 wrist-mounted camera images (left wrist + right wrist) \\
input image size & 224 x 224 px \\
use observation history & yes (2-step history) \\
action chunk size & 24 steps (predict 24, execute all 24 open-loop at test time) \\
use proprio (robot state) & yes \\
image backbone & ResNet-50 \\
\# trainable parameters & 157M \\
diffusion sampling algorithm & DDIM \citep{song2020denoising} \\
number of diffusion steps & 100 at train time; 10 at test time (similar to \citep{chi2023diffusion, khazatsky2024droid}) \\
image augmentations & color jitter, 80\% random crops (as specified by default by \citep{khazatsky2024droid}) \\
\bottomrule
\end{tabular}
\end{table*}

\begin{table*}
\centering
\caption{\textbf{RDT-1B hyperparameters for ALOHA experiments.} This configuration follows default parameters from the RDT-1B codebase \citep{liu2024rdt}.}
\label{tab:rdt_hyperparameters_aloha}
\begin{tabular}{ll}
\toprule
hyperparameter & value \\
\midrule
\# GPUs & 1 x NVIDIA H100 (80GB VRAM) \\
learning rate (LR) & 1e-4 \\
total batch size & 32 \\
\# train steps & 86K for ``fold shorts'' \\
& 30K for ``fold shirt'' \\
& 18K for ``scoop X into bowl'' (performs better than 40K checkpoint) \\
& 150K for ``put X into pot'' \\
input images & 1 third-person camera image, 2 wrist-mounted camera images (left wrist + right wrist) \\
input image size & 256 x 256 px \\
use observation history & yes (2-step history) \\
action chunk size & 25 steps (predict 64 as in \citep{liu2024rdt}, execute 25 open-loop at test time) \\
use proprio (robot state) & yes \\
\# trainable parameters & 1.2B \\
diffusion sampling algorithm & DDPM \citep{ho2020denoising} at train time; DPM-Solver++ \citep{lu2022dpm} at test time \\
number of diffusion steps & 1000 at train time; 5 at test time (same as in \citep{liu2024rdt}) \\
image augmentations & color jitter, image corruption (see \citep{liu2024rdt} for details) \\
\bottomrule
\end{tabular}
\end{table*}

\begin{table*}
\centering
\caption{\textbf{$\pi_0$ hyperparameters for ALOHA experiments.} This configuration follows the default settings specified in the original $\pi_0$ project codebase~\citep{black2024pi_0}.}
\label{tab:pi0_hyperparameters_aloha}
\begin{tabular}{ll}
\toprule
hyperparameter & value \\
\midrule
\# GPUs & 1 x NVIDIA A100 or H100 (80GB VRAM) \\
learning rate (LR) & 2.5e-5 peak LR (1K steps linear warmup, 29K steps cosine decay to 2.5e-6)  \\
total batch size & 32 \\
\# train steps & 115K for ``fold shorts'' \\
& 75K for ``fold shirt'' \\
& 80K for ``scoop X into bowl'' \\
& 80K for ``put X into pot'' \\
input images & 1 third-person camera image, 2 wrist-mounted camera images (left wrist + right wrist) \\
input image size & 224 x 224 px \\
use observation history & no (use single-step inputs) \\
action chunk size & 25 steps (predict 25, execute all 25 open-loop at test time) \\
use proprio (robot state) & yes \\
\# trainable parameters & 3.3B \\
diffusion sampling algorithm & flow matching \citep{lipman2022flow} \\
number of integration steps & 10 \\
image augmentations & random crop (for non-wrist images), random rotation (for non-wrist images), color jitter: \\
& \;\; \texttt{augmax.RandomCrop(int(width * 0.95), int(height * 0.95))} \\
& \;\; \texttt{augmax.Rotate((-5, 5))} \\
& \;\; \texttt{augmax.ColorJitter(brightness=0.3, contrast=0.4, saturation=0.5)} \\
& (same as \cite{black2024pi_0}) \\
\bottomrule
\end{tabular}
\end{table*}

\clearpage

\begin{table*}[h]
\centering
\caption{\textbf{Detailed results for the ``fold shorts'' ALOHA task.} Success is assessed using a staged scoring system, with 20 points allocated to each of five task stages. Points are awarded cumulatively, requiring successful completion of prior stages to receive a nonzero score in subsequent stages. The reported values represent the average performance across all experimental trials.}
\label{tab:detailed_aloha_fold_shorts_results}
\resizebox{\textwidth}{!}{\begin{tabular}{l|c|c|c|c|c|c}
\toprule
& Grasped bottom edge & Folded in half & Grasped waistband  & Folded in half again & Moved shorts forward & Total \\
& (20 pts) & (20 pts) & (20 pts) & (20 pts) & (20 pts) & (100 pts) \\
\midrule
ACT & 20 & 20 & 20 & 20 & 20 & \textbf{100} \\
Diffusion Policy & 20 & 20 & 20 & 20 & 20 & \textbf{100} \\
RDT-1B & 20 & 20 & 20 & 20 & 20 & \textbf{100} \\
$\pi_0$ & 20 & 20 & 20 & 20 & 20 & \textbf{100} \\
OpenVLA-OFT+ & 20 & 20 & 20 & 20 & 20 & \textbf{100} \\
\bottomrule
\end{tabular}}
\end{table*}

\begin{table*}[h]
\centering
\caption{\textbf{Detailed results for the ``fold shirt'' ALOHA task.} Success is assessed using a staged scoring system, with 10 points allocated to each of ten task stages. Points are awarded cumulatively, requiring successful completion of prior stages to receive a nonzero score in subsequent stages. A penalty of -10 points is applied if a large portion of the shirt sticks out after the final fold, as this is a common error for one method and is visually apparent. The reported values represent the average performance across all experimental trials.}
\label{tab:detailed_aloha_fold_shirt_results}
\resizebox{\textwidth}{!}{\begin{tabular}{l|c|c|c|c|c|c|c|c|c|c|c|c}
\toprule
& Grasped bottom edge & Folded in half & Grasped sleeves & Folded sleeves over & Grasped bottom edge & Folded in half & Grasped right edge & Folded in half & Released shirt & Moved shirt forward & Large part sticks out & Total \\
& (10 pts) & (10 pts) & (10 pts) & (10 pts) & (10 pts) & (10 pts) & (10 pts) & (10 pts) & (10 pts) & (10 pts) & (-10 pts) & (100 pts) \\
\midrule
ACT & 10 & 10 & 10 & 10 & 9 & 9 & 9 & 9 & 9 & 9 & 0 & 94 \\
Diffusion Policy & 10 & 10 & 10 & 10 & 10 & 10 & 10 & 10 & 10 & 10 & 0 & \textbf{100} \\
RDT-1B & 10 & 10 & 10 & 10 & 10 & 10 & 10 & 10 & 10 & 10 & -4 & \underline{96} \\
$\pi_0$ & 10 & 10 & 10 & 10 & 10 & 10 & 10 & 10 & 10 & 10 & 0 & \textbf{100} \\
OpenVLA-OFT+ & 10 & 10 & 10 & 10 & 10 & 10 & 10 & 10 & 10 & 10 & 0 & \textbf{100} \\
\bottomrule
\end{tabular}}
\end{table*}

\begin{table*}[h]
\centering
\caption{\textbf{Detailed results for the ``scoop X into bowl'' ALOHA task.} Success is assessed using a staged scoring system. Points are awarded cumulatively, requiring successful completion of prior stages to receive a nonzero score in subsequent stages. Penalties of -5 points are applied if any food is spilled onto the table or if the spoon is not fully emptied during pouring. The reported values represent the average performance across all experimental trials.}
\label{tab:detailed_aloha_scoop_bowl_results}
\resizebox{\textwidth}{!}{\begin{tabular}{l|c|c|c|c|c|c|c|c|c|c}
\toprule
& Moved bowl to center & Grasped spoon & Approached correct container & Scooped correct item & Poured into bowl & Placed spoon next to bowl & Spilled food on table & Did not empty spoon & Total \\
& (10.00 pts) & (10.00 pts) & (20.00 pts) & (20.00 pts) & (20.00 pts) & (20.00 pts) & (-5.00 pts) & (-5.00 pts) & (100.00 pts) \\
\midrule
ACT & 9.17 & 9.17 & 18.33 & 15.00 & 15.00 & 5.00 & -0.42 & 0.00 & 71.25 \\
Diffusion Policy & 10.00 & 10.00 & 18.33 & 15.00 & 13.33 & 13.33 & -0.83 & 0.00 & 79.17 \\
RDT-1B & 7.50 & 7.50 & 15.00 & 15.00 & 15.00 & 13.33 & 0.00 & 0.00 & 73.33 \\
$\pi_0$ & 10.00 & 10.00 & 20.00 & 20.00 & 16.67 & 16.67 & 0.00 & 0.00 & \underline{93.33} \\
OpenVLA-OFT (no FiLM) & 10.00 & 9.17 & 6.67 & 5.00 & 3.33 & 1.67 & -0.83 & 0.00 & 35.00 \\
OpenVLA-OFT+ & 10.00 & 10.00 & 20.00 & 20.00 & 20.00 & 20.00 & 0.00 & 0.00 & \textbf{100.00} \\
\bottomrule
\end{tabular}}
\end{table*}

\begin{table*}[h]
\centering
\caption{\textbf{Detailed results for the ``put X into pot'' ALOHA task (combined results for in-distribution and out-of-distribution evaluations).} Success is assessed using a staged scoring system. Points are awarded cumulatively, requiring successful completion of prior stages to receive a nonzero score in subsequent stages. The reported values represent the average performance across all experimental trials.}
\label{tab:detailed_aloha_put_x_into_pot_results}
\resizebox{\textwidth}{!}{\begin{tabular}{l|c|c|c|c|c|c|c}
\toprule
& Opened pot & Approached correct object & Touched correct object & Grasped correct object & Put correct object into pot & Placed lit on pot & Total \\
& (10.00 pts) & (20.00 pts) & (10.00 pts) & (20.00 pts) & (20.00 pts) & (20.00 pts) & (100.00 pts) \\
\midrule
ACT & 10.00 & 6.67 & 2.08 & 1.67 & 1.67 & 1.67 & 23.75 \\
Diffusion Policy & 10.00 & 7.50 & 3.33 & 3.33 & 3.33 & 3.33 & 30.83 \\
RDT-1B & 10.00 & 11.67 & 5.00 & 5.83 & 5.83 & 5.83 & \underline{44.17} \\
$\pi_0$ & 10.00 & 10.83 & 4.58 & 6.67 & 5.00 & 5.00 & 42.08 \\
OpenVLA-OFT (no FiLM) & 10.00 & 6.67 & 3.33 & 5.00 & 3.33 & 3.33 & 31.67 \\
OpenVLA-OFT+ & 10.00 & 15.83 & 6.25 & 8.33 & 5.83 & 5.00 & \textbf{51.25} \\
\bottomrule
\end{tabular}}
\end{table*}

\begin{table*}[h]
\centering
\caption{\textbf{Additional LIBERO experiments: Single policy for all LIBERO task suites and FiLM ablation study.} Here we train a single policy on all four LIBERO task suites combined, without FiLM (``OpenVLA-OFT'') and with FiLM (``OpenVLA-OFT+''). Policy inputs here include a third-person camera image, wrist camera image, robot proprioceptive state, and a language instruction. We observe comparable task performance when training on all task suites together versus training on each of them independently. In addition, we find that FiLM has little effect on task performance in LIBERO, given that the variant without FiLM already exhibits good language following.}
\label{tab:additional_libero_results}
\begin{tabular}{l|c|c|c|c|c}
\toprule
& Spatial & Object & Goal & Long & Average \\
& SR (\%) & SR (\%) & SR (\%) & SR (\%) & SR (\%) \\
\midrule
OpenVLA-OFT (1 policy per suite) (original, from Table \ref{tab:policy_performance_results}) & 97.6 & \textbf{98.4} & \underline{97.9} & \underline{94.5} & \textbf{97.1}  \\
OpenVLA-OFT (1 policy for all 4 suites) & \underline{97.7} & 98.0 & 96.1 & \textbf{95.3} & 96.8  \\
OpenVLA-OFT+ (1 policy for all 4 suites, + FiLM) & \textbf{97.8} & \underline{98.2} & \textbf{98.2} & 93.8 & \underline{97.0}  \\
\bottomrule
\end{tabular}
\end{table*}

\begin{table*}[h]
\centering
\caption{\textbf{Ablation experiment: Fine-tuning OpenVLA from scratch with the OFT recipe.}  Policy inputs here include a third-person camera image, a wrist camera image, robot proprioceptive state, and a language instruction. The from-scratch policies generally perform worse than the full OpenVLA-OFT policies, confirming that OpenVLA's pretrained representation is beneficial for downstream policy performance even when the fine-tuning recipe differs substantially from the pretraining recipe.}
\label{tab:app:openvla_scratch_ablation_libero}
\begin{tabular}{l|c|c|c|c|c}
\toprule
& Spatial & Object & Goal & Long & Average \\
& SR (\%) & SR (\%) & SR (\%) & SR (\%) & SR (\%) \\
\midrule
OpenVLA-OFT & 97.6 & 98.4 & 97.9 & 94.5 & 97.1  \\
OpenVLA-OFT (scratch) & 94.3 & 95.2 & 91.7 & 86.5 & 91.9 \\
\bottomrule
\end{tabular}
\end{table*}

\begin{table*}[h]
\centering
\caption{\textbf{OpenVLA-OFT in BridgeData V2 WidowX robot evaluations.} We use a subset of the tasks from the original OpenVLA BridgeData V2 evaluation suite, along with the corresponding scoring criteria. Each score is the average over 10 trials per task. OpenVLA-OFT surpasses OpenVLA in average task performance and even demonstrates effective language following without FiLM.}
\label{tab:bridge_results}
\begin{tabular}{llcc}
\toprule
Category & Task & OpenVLA Score & OpenVLA-OFT Score \\
\midrule
Visual gen & Put Eggplant into Pot & 60 & \textbf{90} \\
Motion gen & Lift Eggplant & \textbf{70} & 60 \\
Physical gen & Flip Pot Upright & \textbf{90} & 80 \\
Semantic gen & Lift White Tape & 0 & \textbf{20} \\
Language grounding & Put \{Blue Cup, Pink Cup\} on Plate & 85 & \textbf{90} \\
Language grounding & Lift \{Cheese, Red Chili Pepper\} & \textbf{90} & 75 \\
\midrule
& Average & 65.8 & \textbf{69.2} \\
\bottomrule
\end{tabular}
\end{table*}

\end{document}